\documentclass{article}

\usepackage{microtype}
\usepackage{graphicx}
\usepackage{subcaption}
\usepackage{booktabs} 

\usepackage{hyperref}

\usepackage[preprint]{icml2026}

\usepackage{amsmath}
\usepackage{amssymb}
\usepackage{mathtools}
\usepackage{amsthm}

\usepackage{amsmath,amsfonts,bm}

\def\figref#1{figure~\ref{#1}}

\def\twofigref#1#2{figures \ref{#1} and \ref{#2}}

\def\secref#1{section~\ref{#1}}

\def\eqref#1{equation~\ref{#1}}

\def\1{\bm{1}}

\DeclareMathAlphabet{\mathsfit}{\encodingdefault}{\sfdefault}{m}{sl}
\SetMathAlphabet{\mathsfit}{bold}{\encodingdefault}{\sfdefault}{bx}{n}

\newcommand{\R}{\mathbb{R}}

\DeclareMathOperator{\floateq}{\dot{=}}
\def\supfigref#1{supplemental figure~\ref{#1}}
\def\twosupfigref#1#2{supplemental figures~\ref{#1} and~\ref{#2}}
\def\threesupfigref#1#2#3{supplemental figures~\ref{#1}, \ref{#2}, and~\ref{#3}}
\def\appendixref#1{appendix~\ref{#1}}

\usepackage[capitalize,noabbrev]{cleveref}

\theoremstyle{plain}

\theoremstyle{definition}

\theoremstyle{remark}

\usepackage[textsize=tiny]{todonotes}

\icmltitlerunning{Delayed Robustness Is Not Grokking}

\begin{document}

\twocolumn[
  \icmltitle{Delayed Robustness Is Not Grokking}



  \icmlsetsymbol{equal}{*}

  \begin{icmlauthorlist}
    \icmlauthor{Stijn van den Beemt}{yyy}
    \icmlauthor{Twan van Laarhoven}{yyy}
    \icmlauthor{Marco Loog}{yyy}
  \end{icmlauthorlist}

  \icmlaffiliation{yyy}{iCIS Institue for Computer and Information Sciences, Department of Data Science (DaS), Radboud University, Nijmegen, The Netherlands}

  \icmlcorrespondingauthor{Stijn van den Beemt}{stijn.vandenbeemt@ru.nl}

  \icmlkeywords{neural networks, delayed robustness, grokking, adversarial attacks, deep learning}

  \vskip 0.3in
]



\printAffiliationsAndNotice{}  

\begin{abstract}
    We analyze the phenomenon of delayed robustness, where a neural network trained beyond overfitting becomes robust to adversarial attacks. This phenomenon was first observed by \citet{ALLNETWORKSGROK}, and characterized as grokking. We reproduce delayed robustness to PGD attacks in multiple set-ups and show that delayed robustness is the result of weak attacks and two unintended side effects when overtraining: 1) softmax collapse in the cross-entropy loss function, and 2) a too large effective learning rate caused by gradient scaling in the Adam optimizer. We demonstrate that these issues create networks resistant to PGD attacks but still vulnerable to stronger attacks. We also point out an interesting relation between delayed robustness, the slingshot effect, and dead neurons. Using interventions to solve the softmax collapse and gradient scaling issues, we show that no delayed robustness emerges. We must therefore conclude that delayed robustness is not grokking.
\end{abstract}

\section{Introduction and Background}
	Grokking, where an over-parameterized neural network starts to generalize well
	past the point of overfitting, has been an active object of deep learning
	research ever since it was first described by \citet{GROKKINGPOWER}. This
	includes research on relevant training settings, like the importance of weight
	decay \citep{OMNIGROK} and the design of new optimization techniques to
	accelerate grokking behavior \citep{lee2024grokfast}. This training past
	overfitting also uncovers other phenomena that occur late into training, like
	delayed robustness from \citet{ALLNETWORKSGROK}. They show how ReLU networks
	consistently become robust to adversarial attacks after overtraining. Their
	hypothesis is that networks undergo a phase transition that
	creates robust partitions of input space, which is why they called this grokking of adversarial examples.
	We want to further test this hypothesis.
	Our main goal is to determine whether or not this effect can truly
	be considered grokking.

	In \secref{DRsection}, we discuss what it means for a network to be robust
	against adversarial attacks and how to accurately determine such robustness. We
	reproduce and extend the findings of \citet{ALLNETWORKSGROK} and show that these
	networks are not as robust as they seem. This leads to an alternative hypothesis described in \secref{gradientkaputsec}: that delayed robustness can be attributed to unintended side effects caused by overtraining.
	In particular, we look at softmax collapse in the cross-entropy loss degenerating
	the gradient, and also at issues related to non-convergent behavior of the
	gradient scaling mechanism in the Adam optimizer. In \secref{MLPex}, we share	results of experiments in multiple architectures, that, as argued in the discussion in \secref{discussion} show that it is indeed these issues, and not grokking, that underlie the delayed robustness phenomenon.

	\section{Preliminaries: (Delayed) Robustness}\label{DRsection}
	We start by explaining adversarial robustness and subsequently define and reproduce the
	delayed robustness phenomenon.

	\subsection{What is a Robust Network?}
	A neural network, like any classifier, should in theory be invariant, or
	robust, to small and irrelevant changes in the input when making a
	classification decision. This quality has been a topic in ML research for over
	two decades \citep[e.g.][]{dalvi2004adversarial}. Formally, a classification
	neural network $f$ is robust with respect to a data sample $x$ with dimension $N$, and with true label $y$, to a perturbation $\epsilon$ measured by a
	distance norm $d$ if and only if:
	\begin{align}
		\forall x' \in \R^N: |x - x'|_d  \leq \epsilon \Rightarrow f(x') = y
		\label{rob-crit}.
	\end{align}
	An adversarial attack is a strategy to modify a sample such that it crosses a
	decision boundary into a wrong class within such a perturbation. The percentage
	of input samples still correctly classified after perturbation by such an attack
	is then a measure of robustness, referred to here as adversarial accuracy but
	also known as robust accuracy \citep{AUTOATTACK}.

	\subsection{Delayed Robustness}\label{DRreprod}
\begin{figure*}[h]
		\begin{center}
			\includegraphics[width=\linewidth]{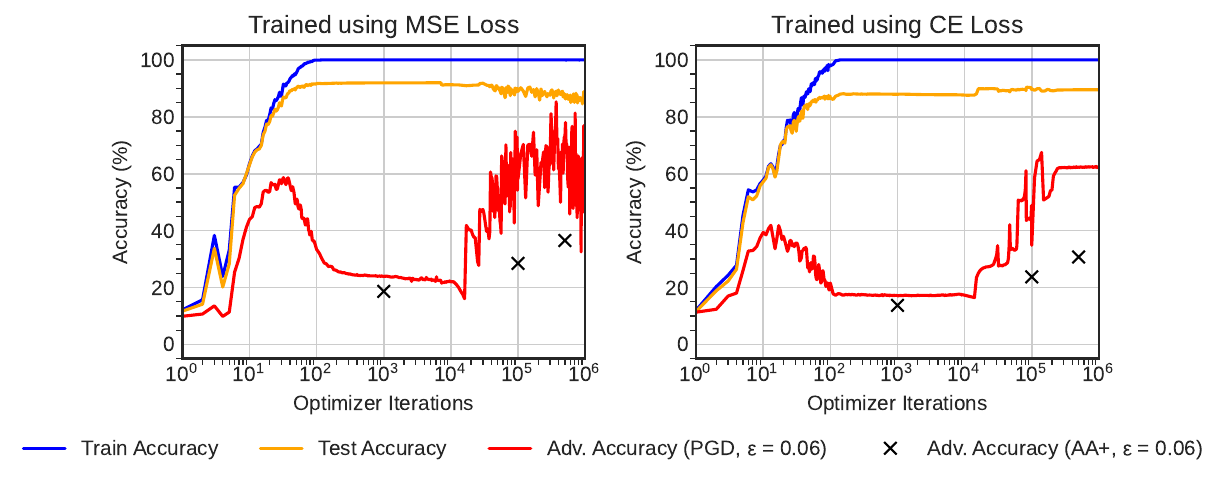}
		\end{center}
		\caption{\textit{Train, test and adversarial accuracies (from perturbing the
				test set using $l_\infty$-PGD with and AA+ $\epsilon = 0.06$) of two MNIST-trained MLPs
				trained with mean squared error loss (left) and softmax cross-entropy loss
				(right) respectively.}}\label{DRfig}
	\end{figure*}

	In \figref{DRfig}, we show what \citet{ALLNETWORKSGROK} called \emph{delayed robustness}: a sudden increase in adversarial robustness late in training, long after both train- and test accuracy convergence. We train an MLP network on
	1000 random MNIST dataset \citep{MNIST} samples for a large number of batch
	iterations with the ADAM optimizer.\footnote{Whenever we mention training in this paper, we mean standard training and not adversarial training.} We have two setups: one uses a mean squared
	error (MSE) loss function and the other softmax cross-entropy (CE) loss.
	We then used an $\infty$-norm PGD (Projected Gradient Descent, \citealp{PGD})
	attack on the full test set with $\epsilon = 0.06$. This figure reproduces and extents results of \citet{ALLNETWORKSGROK}. Full experimental details
	are in \appendixref{expdetails}.

	\citet{ALLNETWORKSGROK} attributed this delayed robustness to the network creating a more robust partition of the input space,
	and claimed that it is a form of grokking.
	Their argumentation is the following: A network with piecewise-linear activations is itself also a piecewise-linear function. Such a network partitions the input space into linear regions, each defined by a unique set of neurons that have positive pre-activations.
	This concept led to their measure of local complexity: the number of such regions near a
	data point. They showed a decrease of local complexity around training
	data points and ascribed this to region migration, a phenomenon where
	these regions move away from the training data and close(r) to the decision
	boundaries. This region migration is their hypothesized driver for the robust partition and thus the supposed grokking. We refer to their paper for the exact details.

	However, when we attack these networks with an attack stronger than PGD,
	like the parameter-free AutoAttack+ (AA+, \citealp{AUTOATTACK}), the networks
	show much less adversarial accuracy, which is also illustrated in \figref{DRfig}. As
	the amount of perturbation $\epsilon$ increases, the remaining adversarial
	accuracy decreases; see \supfigref{PGDvsAAfig}. 

	\section{A New Hypothesis for Delayed Robustness}\label{gradientkaputsec}
	The results in \secref{DRsection} show that using PGD, adversarial robustness is overestimated,
	especially at higher $\epsilon$.
	We therefore propose an alternative hypothesis for delayed robustness: since the PGD strategy is to use the gradients of the loss over the input space to maximize the cross-entropy loss \citep{PGD}, delayed
	robustness to PGD can be explained by the computed gradient becoming uninformative for PGD attacks when overtraining.

	We will explain two different issues that could cause these effects and also mention solutions that we apply in our experiments.

	\subsection{Gradient Scaling in ADAM}\label{gradscalesection}
	The first issue relates to the Adam optimizer \citep{kingma2014adam}, which
	uses gradient scaling to adapt the effective learning rate relative to the
	gradients' magnitudes. At step $t$, a network $f$ with parameters $\theta$ and
	true labels $y$ over input $x$ is optimized using the following update
	rules:
	\begin{align}
		g_t & \gets \nabla_\theta \ell(f(x; \theta_{t-1}), y)\\
		m_t & \gets \beta_1m_{t-1} + (1-\beta_1)g_t\\
		v_t & \gets \beta_2v_{t-1} + (1-\beta_2)g_t^2\\
		\theta_t & \gets \theta_{t-1} - \frac{\gamma}{\sqrt{v_t} + \delta}
		m_t\label{gradscale}.
	\end{align}
	Here $\gamma$ is the base learning rate and $\delta$ a small value added to
	avoid division by zero.\footnote{Other work often uses $\epsilon$ to denote this
	small value; here we use $\delta$ to avoid confusion with the perturbation
	$\epsilon$ from adversarial attacks.} $\beta_1$ and $\beta_2$ are parameters for
	the decay rate of the running averages, with default values of $0.9$ and $0.999$ respectively. Note that we omit the bias correction factors $1/(1-\beta_1^t)$ and $1/(1-\beta_2^t)$, because in an overtraining regime $t$ is sufficiently large, such that they are practically equal to 1.

	When the gradient $g_t$ becomes very small for many steps, so does $v_t$. This
	causes the scale factor $\frac{\gamma}{\sqrt{v_t} + \delta}$ to increase, and as
	a result the effective learning rate can become too large.
	This can cause issues where Adam fails to converge, which has been described
	before by \citet{AMSGRAD}.
	Since $\delta$ acts as an upper bound to the scaling factor, a large enough
	value of $\delta$ prevents this over-amplification of small gradients.
	However, in practice, a very small value $\delta=10^{-8}$ is typically used.
	As \citeauthor{AMSGRAD} note: \textit{In practice, selection of the [$\delta$]
		parameter appears to be critical for the performance of the algorithm.}

	One solution is then to increase the $\delta$ term, thereby lowering the upper
	bound on the effective learning rate.
	Another solution is used in the AMSGrad optimizer proposed by \citeauthor{AMSGRAD}.
	This algorithm keeps track of a maximum for $v$ throughout training, providing
	an alternative upper bound. This algorithm thus replaces \eqref{gradscale} by
	\begin{align}
		v_t^\text{max} & \gets \max(v_{t-1}^\text{max}, v_{t-1})\\
		\theta_t & \gets \theta_{t-1} - \frac{\gamma}{\sqrt{v_t^\text{max}} + \delta}
		m_{t} \label{AMSGrad}.
	\end{align}

	\subsection{Softmax Collapse}\label{SMCsection}
	A second issue that occurs during overtraining is softmax
	collapse, which is caused by numerical precision errors in calculating the
	softmax cross-entropy loss function or its gradient.
	The term was introduced by \citet{SOFTMAXCOLLAPSE}, who show that the
	exponential part of cross-entropy loss is at risk of floating-point precision
	errors, even when written in the more stable form of
	\begin{align}
		\ell_{SCE}(\textbf{z}, y) = -z_y + \max(\textbf{z}) + \log
		\biggl(\sum_{k=1}^{n} e^{z_k - \max(\textbf{z})}\biggr)\label{SCE},
	\end{align}
	where $\textbf{z}$ are the classification network's $n$ output logits, $y$ is
	the true class label, and so $z_y$ corresponds to the output logit
	for to the correct class.

	We will refer to the rightmost term of \eqref{SCE} as the
	\texttt{logsumexp} term, after the equivalent PyTorch
	function\footnote{\url{https://docs.pytorch.org/docs/stable/generated/torch.logsumexp.html}}. We discern two types of collapse that both involve this term: underflow softmax collapse and absorption softmax collapse.

	\subsubsection*{Underflow Softmax Collapse}
	When the difference between an incorrect
	logit and the maximum logit becomes so large that $e^{z_k - \max(\textbf{z})}$
	does not fit within the numerical range of the float representation anymore, it
	evaluates to 0 (underflow). If none of the incorrect logit differences fit
	anymore, only the term from the correct logit $z_y = \max(\textbf{z})$ is left,
	and the whole \texttt{logsumexp} returns 0. Because
	\begin{align}
		\frac{\partial\ell_{SCE}}{\partial z_c} = \frac{e^{z_c -
				\max(\mathbf{z})}}{\sum^n_{k=1}e^{z_k - \max(\mathbf{z})}} - \mathbf{1}_{c =
			y},\label{SCEGrad}
	\end{align}
	the gradients with respect to underflowed logits also underflow, while
	increasing the gradients with respect to the remaining logits. This phenomenon
	was also called gradient vanishing, and first described by
	\citet{carlini2017towards}. We however refer to this as underflow softmax collapse, to distinguish it from the other type of collapse and avoid confusion with other gradient vanishing phenomena.

	\subsubsection*{Absorption Softmax Collapse}
	A second type of softmax collapse, absorption softmax collapse, was described
	by \citet{SOFTMAXCOLLAPSE}. Absorption occurs when adding a small float to a much larger float, and the result is rounded to the largest float. We refer to \citet{SOFTMAXCOLLAPSE} for a more detailed description. They note that the \texttt{logsumexp} term can also
	reduce to 0 for a correctly predicted sample through such floating-point absorption,
	\begin{align}
		\sum_{k=1}^{n} e^{z_k - \max(\mathbf{z})} \floateq e^0 = 1\label{FPE},
	\end{align}
	where $\floateq$ denotes floating-point equality.

	This type of softmax collapse can still occur even when all the exponential
	terms fit within float precision individually. Also, the gradient of this type
	of softmax collapse is only 0 for the correct class logit, while the gradients
	of others become equivalent to their exponential term, as demonstrated by
	combining \eqref{SCEGrad} and \eqref{FPE}.

	\subsubsection*{Solutions}\label{SCSolutions}
	A band-aid solution to both types of softmax collapse is to simply increase the
	numerical precision of the calculations. However, any finite-precision calculation is a temporary solution, since logits keep growing using cross-entropy loss
	through a process called naive loss minimization \citep{SOFTMAXCOLLAPSE}. Note
	that for absorption softmax collapse, it is sufficient to increase the precision
	only when calculating the loss and the logit gradients from the network's output
	logits, before back-propagation. For underflow softmax collapse, it is also
	necessary to increase the precision of the back-propagation.

	Another solution is to scale the logits to an input domain where the
	calculation is numerically stable. This results in a scaled variant of softmax
	cross-entropy loss (ScaledSCE),
	\begin{align}
		\ell_{ScaledSCE}(\mathbf{z}) =
		\ell_{SCE}(s(\mathbf{z}) \mathbf{z} ), \\
		s(\mathbf{z}) = \min\left(1, \frac{s_\text{max}}{\max_{j,k}(z_j -
			z_k)}\right)
		,
	\end{align}
	where $s_\text{max} = 20$ is the maximum allowed difference between logits.
	This loss function can be used as a drop-in replacement for cross-entropy loss in a
	PGD attack. This rescale-and-attack strategy, which we call ScaledPGD, can be
	used to adversarially attack samples where normal PGD fails because of softmax
	collapse. We only use ScaledSCE for attacking and not for training because it does not actually inhibit naive loss minimization during training, which we show in \appendixref{AppendixScaledCE}.

	\section{Experiments}\label{MLPex}
	We perform a number of experiments to show that Adam convergence issues and softmax collapse underlie delayed robustness to PGD. We perform most of these
	experiments with MNIST-trained MLPs, but we also validate our results on a
	CIFAR10-trained ResNet18. We refer to \appendixref{expdetails} for
	further experimental details.

	\subsection{Mean Squared Error}\label{resMNISTMSE}
	For our first experiments, we train a four-layer MLP with the same
	initialization on the same MNIST subset of 1000 samples for $10^7$ optimizer
	iterations with three different optimizers. Two runs use the Adam optimizer, one
	with the default $\delta$ value of $10^{-8}$ and another with a value
	$\delta=10^{-4}$. The third run uses the AMSGrad optimizer with the default
	$\delta=10^{-8}$.

\begin{figure*}[h]
		\begin{center}
			\includegraphics[width=1.0\linewidth]{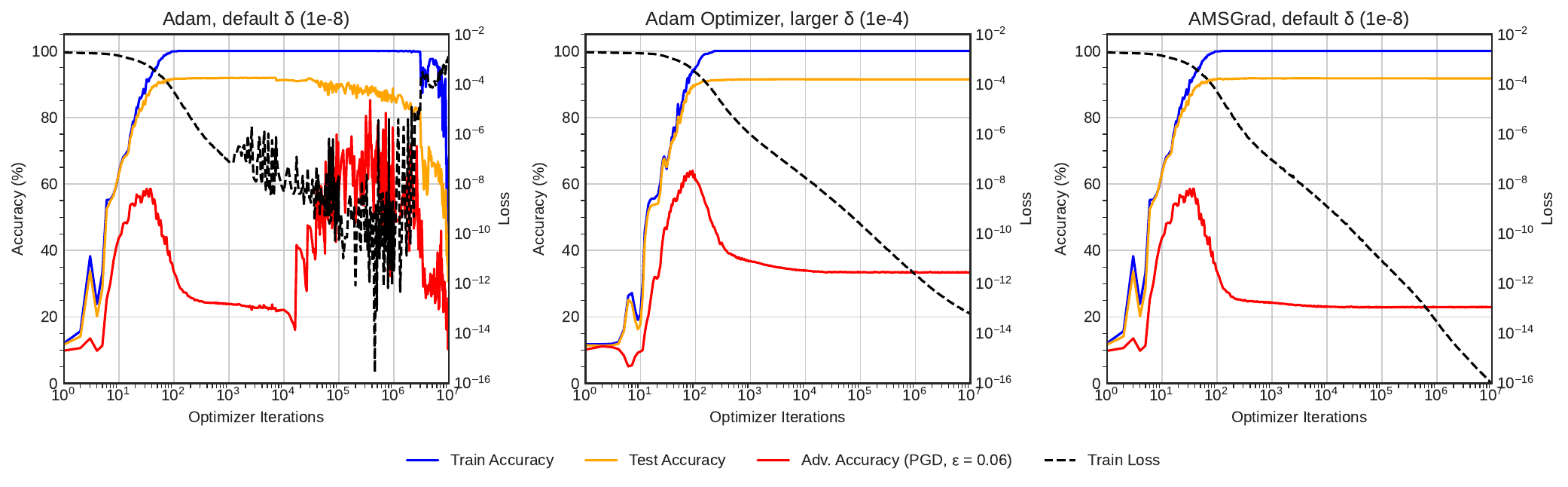}
		\end{center}
		\caption{\textit{Train, test, and adversarial accuracies (from perturbing the
				test set using $l_\infty$-PGD with $\epsilon = 0.06$), as well as train set loss
				of three MNIST-trained MLPs trained with mean squared error loss. One uses the
				default Adam optimizer (left), another Adam with a higher $\delta = 10^-4$
				(middle) and the last one uses the AMSGrad optimizer (right).}}\label{MSEMnistfig}
	\end{figure*}

	For each run, we track the train and test accuracy as well as the train set
	loss and the adversarial accuracy on the test set using PGD. More details can be
	found in the \appendixref{expdetails}, and the results are in
	\figref{MSEMnistfig}. Overtraining using Adam with default parameters and mean
	squared error loss shows delayed robustness to PGD, but also deterioration of
	the network's accuracy and a very unstable loss. The two other optimizers
	clearly stabilize the loss and also do not cause delayed robustness to PGD.

\begin{figure*}[h]
		\begin{center}
			\includegraphics[width=1\linewidth]{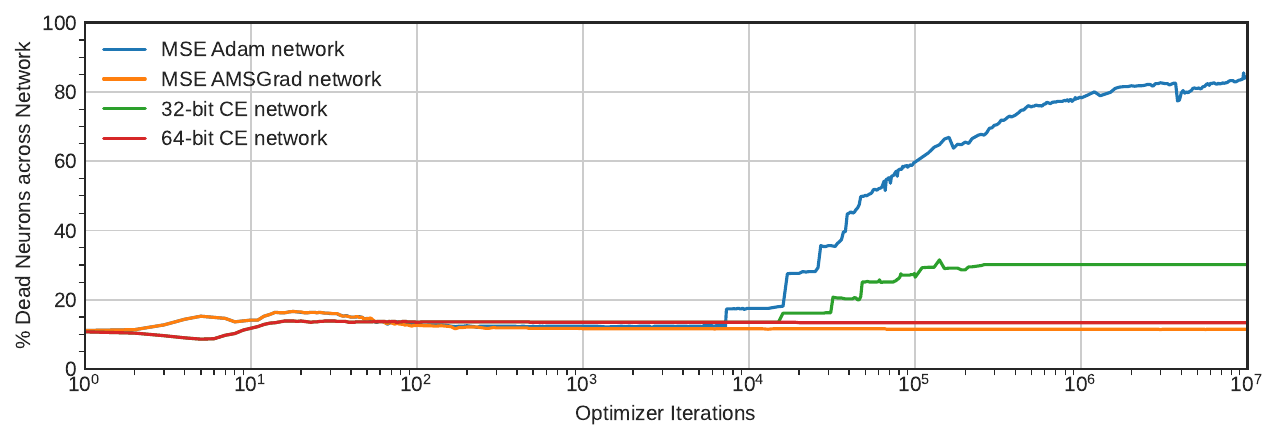}
		\end{center}
		\caption{\textit{Analysis of the number of dead neurons throughout training of
				the MLPs on 1000 MNIST samples in our experimental set-ups.}}\label{MNIST_dead_neurons_graph}
	\end{figure*}

	To understand why the first network ends up in such a poorly performing
	state, we analyzed both the number of dead neurons as well as the local
	complexity, calculated as per \citet{ALLNETWORKSGROK}. We consider a ReLU neuron
	dead if it has 0 post-activation for the entire training data set. The results
	can be found in \twofigref{MNIST_dead_neurons_graph}{MNIST_LC_graph}. They show
	that using Adam with $\delta=10^{-8}$, the local complexity starts decreasing for
	training samples, very similar to \citeauthor{ALLNETWORKSGROK}'s figure~2, around the
	same time that the loss becomes unstable. Subsequently, dead neurons start
	accumulating throughout the layers. In the network trained with AMSGrad,
	however, the local complexity for the train samples rises smoothly before
	evening out, while the number of dead neurons remains stable.

	\begin{figure*}[h]
		\begin{center}
			\includegraphics[width=1\linewidth]{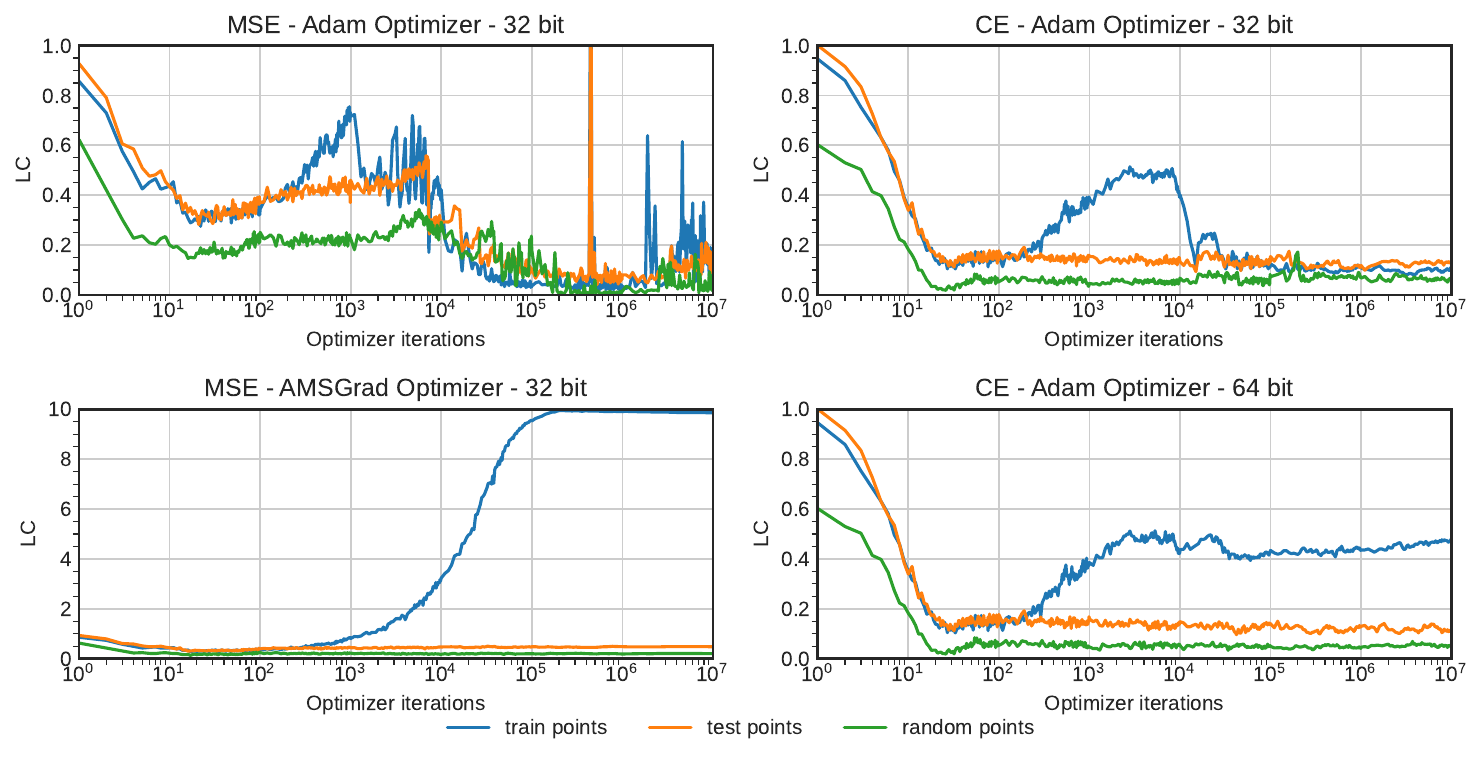}
		\end{center}
		\caption{\textit{Local Complexity (LC), around train, test, and
				random data points, throughout training of the MLPs on 1000 MNIST samples.}}\label{MNIST_LC_graph}
	\end{figure*}

	\subsection{Cross-Entropy}\label{secCEMnist}
\begin{figure*}[h]
		\begin{center}
			\includegraphics[width=0.925\linewidth]{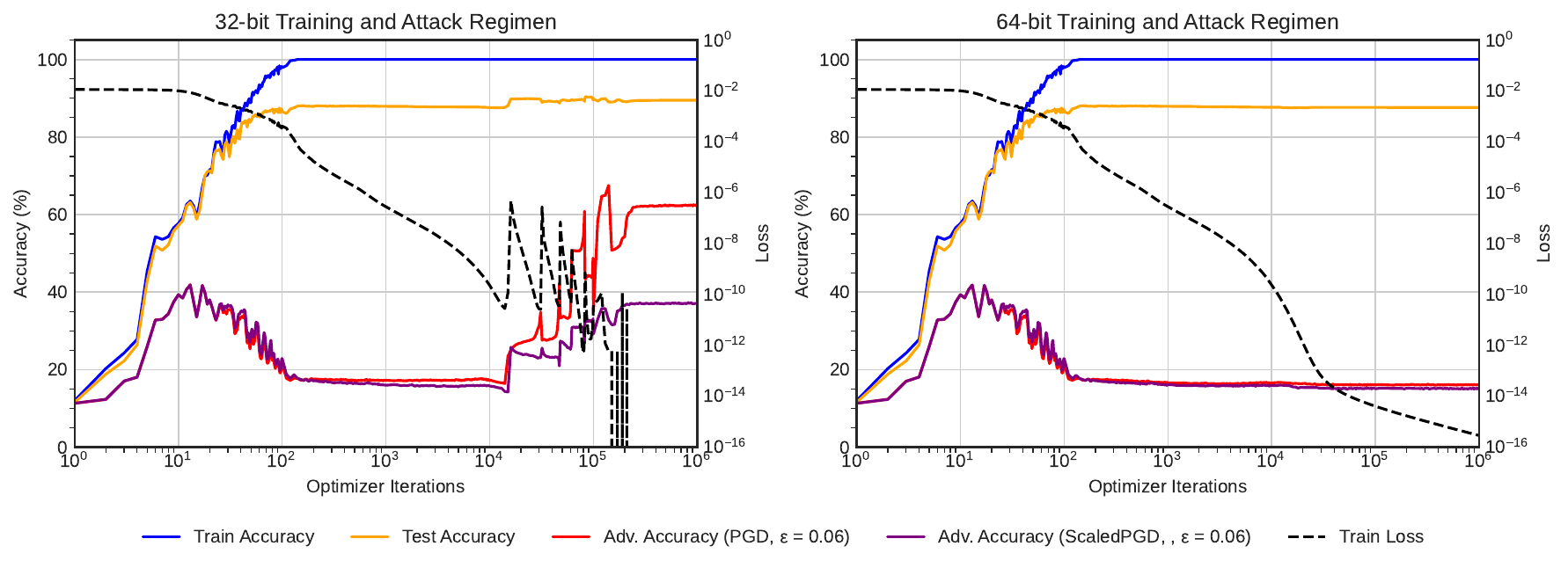}
		\end{center}
		\caption{\textit{Train, test, and adversarial accuracies (from perturbing the
				test set using $l_\infty$-PGD at both 32- and 64-bit precision and
				$l_\infty$-ScaledPGD) with $\epsilon = 0.06$), as well as train set loss of two
				MNIST-trained MLPs trained with softmax cross-entropy loss. One was trained
				using default 32-bit loss calculation (left), the other used a 64-bit loss
				calculation.}}\label{CEMnistfig}
	\end{figure*}

	Our next experiments also use a four-layer MLP, but this time trained using cross-entropy loss for 1\,000\,000 optimizer iterations.
	We train and attack the network twice, once using 32-bit precision before,
	and once using 64-bit precision. We track the same metrics as for the mean squared error networks, as well as the
	adversarial accuracy obtained by ScaledPGD. The results are in
	\figref{CEMnistfig}.
	\begin{figure*}[h]
		\begin{center}
			\includegraphics[width=0.925\linewidth]{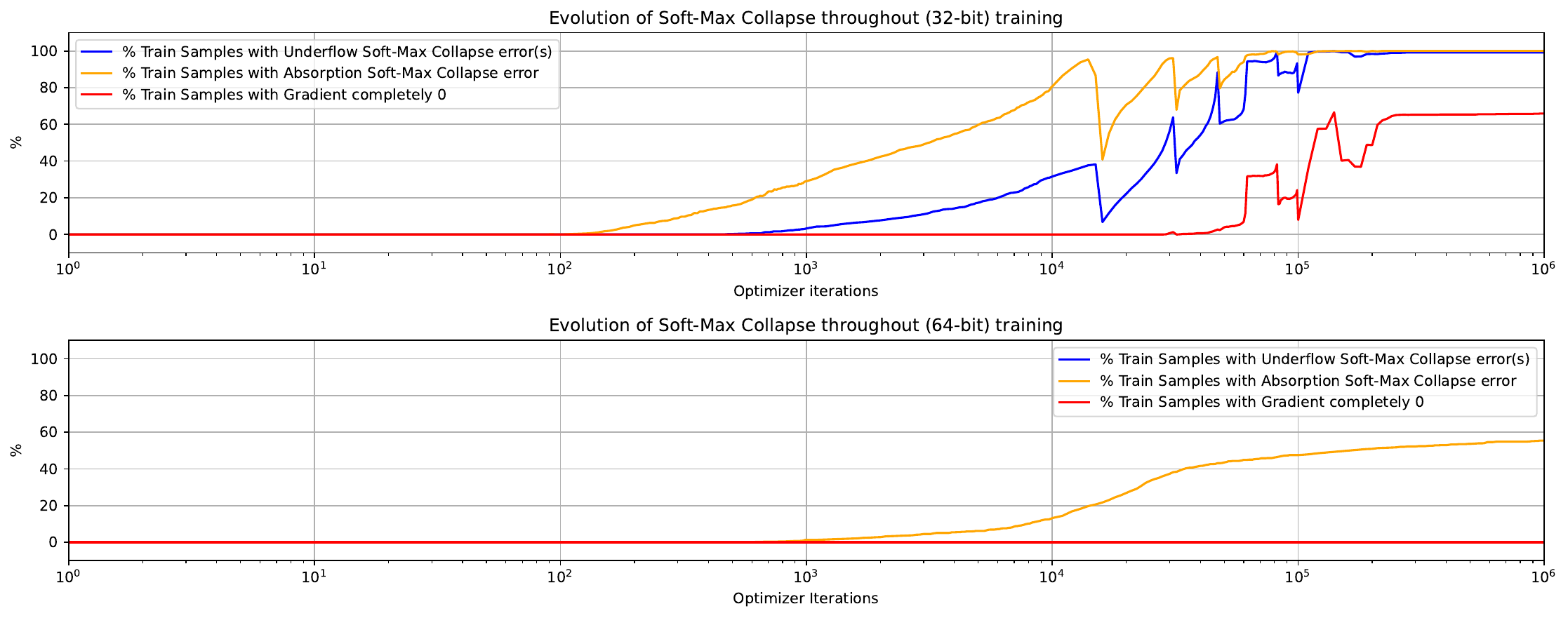}
		\end{center}
		\caption{\textit{Softmax collapse errors in the output logit gradients
				throughout training of the MLPs on 1000 MNIST samples using 32 bits (top) or
				64 bits (bottom) precision.}}\label{CE_MNIST_collapse_graph}
	\end{figure*}

	In 32 bits, we again see an unstable training loss. However, this time these
	instabilities did not affect the train accuracy and instead marginally improved
	our test accuracy.
	Using ScaledPGD, we obtain a far lower adversarial accuracy, just like using AA+
	did in \figref{DRfig}. In \supfigref{CEpgdsfig} we also show the difference
	between default PGD and ScaledPGD attack accuracies at different perturbation
	levels of this network. When training with 64-bit precision, delayed robustness is completely absent
	and all metrics are stable.

	To show that it is softmax collapse underlying these instabilities in the loss,
	we count the occurrences of softmax collapse during training. These results can
	be found in \figref{CE_MNIST_collapse_graph}, which shows that in the 32-bit
	network there is a high amount of softmax collapse when the loss instabilities
	occur. With 64 bits there is only a steadily increasing amount of absorption
	softmax collapse.

	In \twofigref{MNIST_dead_neurons_graph}{MNIST_LC_graph} we show the number of
	dead neurons and the local complexity in these experiments. With 32-bit
	precision, there are sharp increases in the number of dead neurons, co-occurring
	with the loss instabilities. This is preceded by a sharp drop in local
	complexity around training samples. In the 64-bit network, both the local complexity and the number of dead neurons are stable.
	Note that using LeakyReLU does not alleviate the neurons dying (\threesupfigref{MSEMnistfigleaky}{CEMnistfigleaky}{MNIST_dead_neurons_graph_leaky}), as they can still get into a permanently negative output state.

	\subsection{CIFAR10 ResNet18 Validation}\label{resnetsubsec}
	To show that our results generalize to different architectures and datasets, we
	also perform experiments on a ResNet18 \citep{RESNET} architecture. We trained
	the network on the the full 50\,000 sample CIFAR10 dataset \citep{CIFAR10} for 1\,000\,000 optimizer iterations. Like \citet{ALLNETWORKSGROK} we do not use batch
	normalization. We again evaluate the difference between the default Adam
	optimizer and the AMSGrad optimizer when using mean squared error loss, and the
	difference between 32- and 64-bit training when using cross-entropy loss. For
	these experiments, we report adversarial accuracies with respect to PGD, as well as AA+ (for mean squared error loss) and ScaledPGD (for cross-entropy
	loss).All these attacks were done with $\epsilon = 0.06$. The results are in
	\figref{Resnet_adv_accs}. Plots with more metrics are in the appendix (\twofigref{MSECifar10Fig}{CEresnetfig}).
	\begin{figure*}[h]
		\begin{center}
			\includegraphics[width=0.9\linewidth]{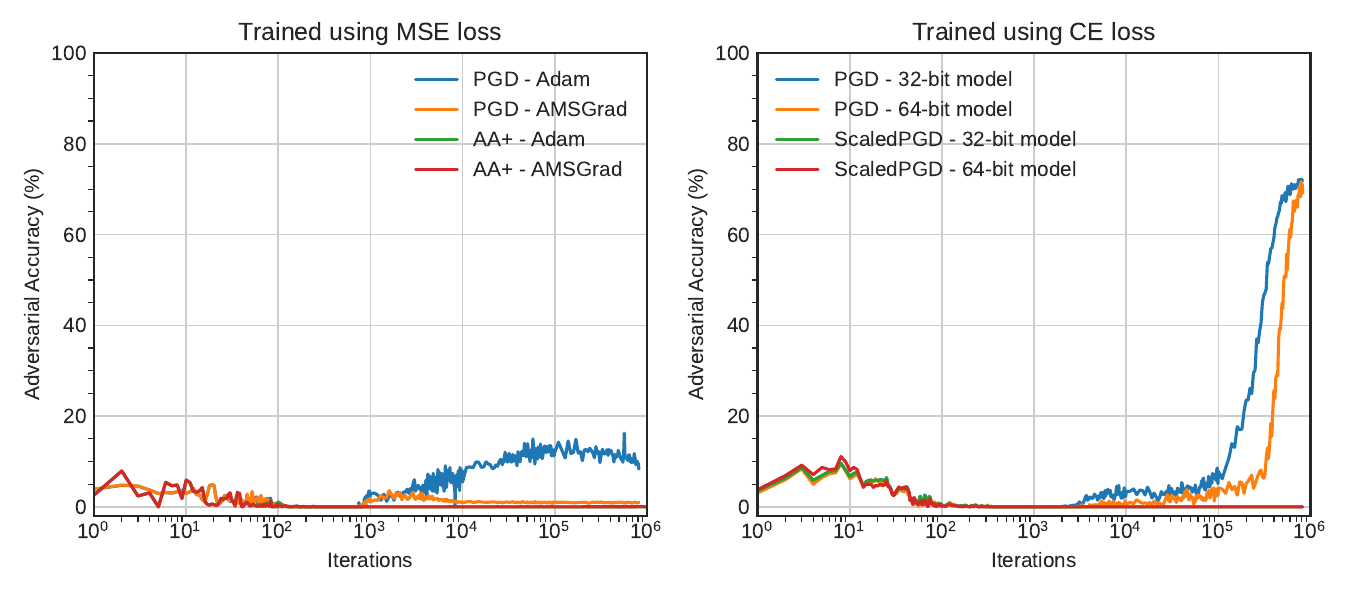}
		\end{center}
		\caption{\textit{Adversarial accuracies to different attacks in the different ResNet18 setups.
				$\epsilon = 0.06$.}}\label{Resnet_adv_accs}
	\end{figure*}

\begin{figure*}[h]
		\begin{center}
			\includegraphics[width=0.9\linewidth]{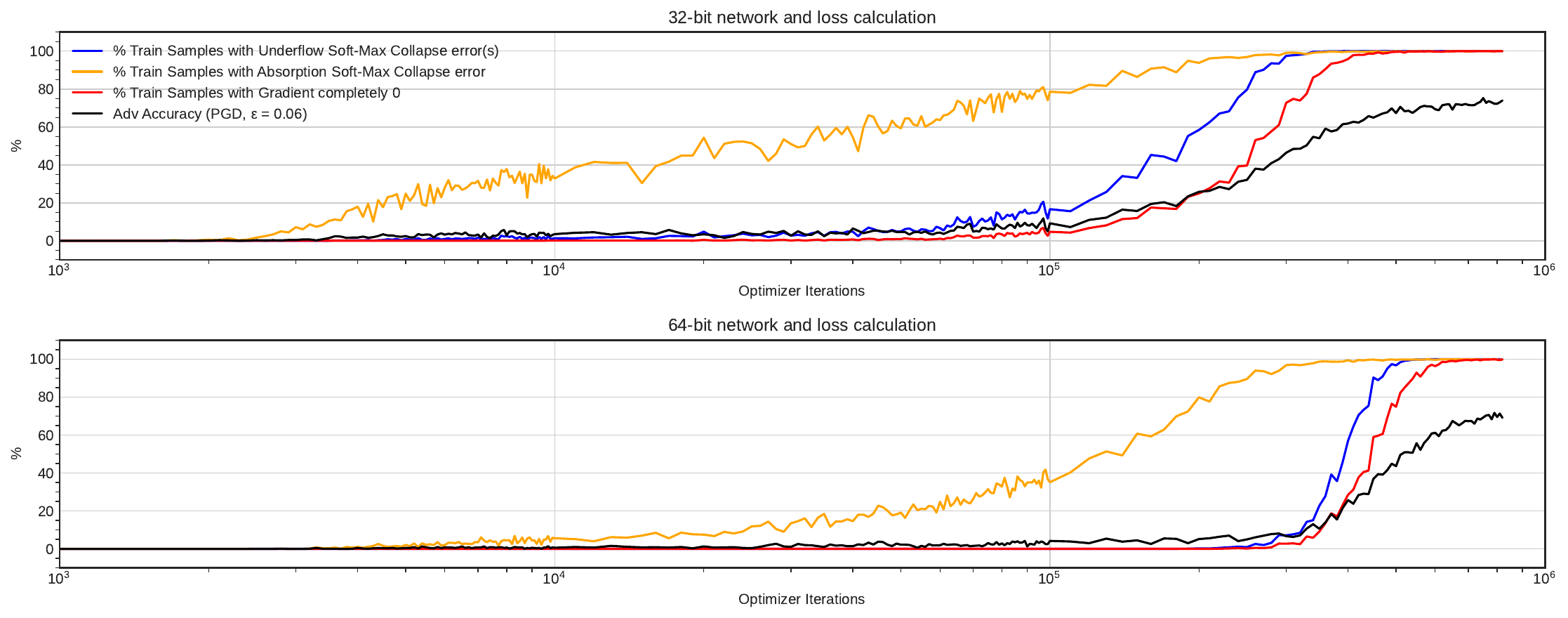}
		\end{center}
		\caption{\textit{Softmax collapse errors in the output logit gradients
				throughout training of the ResNet on CIFAR10 using 32-bit (top) or 64-bit
				(bottom) precision. Also includes the adversarial accuracy against PGD attacks
				on the test set at $\epsilon = 0.06$. Please note
				the x-scale starting at $10^3$ iterations.}}\label{CE_Cifar_collapse_graph}
	\end{figure*}

	The adversarial accuracy results show delayed robustness to PGD in all the
	experiments.
	The mean squared error network trained with Adam shows a small and noisy gain in adversarial accuracy when attacked by PGD. Using AMSGrad further decreased and stabilized this adversarial accuracy.
	The cross-entropy models' results instead show large sudden gains in adversarial accuracy to PGD. This
	closely resembles the pattern of the ResNet experiments of
	\citet{ALLNETWORKSGROK}. Increasing the precision from 32 to 64 bits only delayed
	this pattern, rather than fully preventing it. We show a similar delay in the
	development of softmax collapse errors in \figref{CE_Cifar_collapse_graph}.
	Contrary to all these cases, the AA+ and ScaledPGD attacks leave practically no adversarial accuracy.

	\section{Discussion}\label{discussion}
	The theoretical and experimental evidence presented shows that the delayed
	robustness phenomenon reported by \citet{ALLNETWORKSGROK} is caused by
	unintended side effects during overtraining.

	Using mean squared error loss and
	the Adam optimizer, we managed to precisely reproduce \citeauthor{ALLNETWORKSGROK}'s results on MNIST (their figure 2). We saw delayed robustness against PGD attacks, but also observed an unstable loss and an accumulation of dead neurons, which ultimately harmed the network's performance. Our results indicate that all of these effects are due to Adam's gradient scaling causing a too high effective learning rate. Using AMSGrad
	prevented the loss instabilities, stopped neurons from dying, stabilized
	local complexity and most importantly, did not exhibit delayed robustness. We show we can reproduce these findings using different random subsets of 1000 MNIST samples and other model
	initializations in \twosupfigref{MNIST_repeat_accs_model_MSE}{MNIST_repeat_accs_data_MSE} respectively.

	In the 32-bit MNIST cross-entropy experiments, we also saw delayed robustness using PGD
	and even some gain in adversarial accuracy using ScaledPGD. These gains were also
	associated with loss instabilities, a drop in local complexity, and dying
	neurons. Here, these effects co-occur at some critical threshold of
	softmax collapse. We show that these effects happen consistently in
	\twosupfigref{MNIST_repeat_accs_CE}{MNIST_repeat_grads_CE}, where we repeat these
	findings, again using different seeds. By increasing
	the model's precision to 64 bits, we stopped the majority of collapse and
	thereby also all the subsequent effects, which confirms that numerical errors were their cause. We again also saw no delayed robustness with 64 bit precision.

	We also showed that ResNets are not actually becoming robust when overtrained. In ResNets trained on
	CIFAR10 using mean squared error and Adam, AA+ was still fully
	effective. ResNets trained
	using cross-entropy loss showed delayed robustness to PGD, but were still fully
	vulnerable to ScaledPGD. This delayed robustness can therefore be attributed purely to softmax collapse errors.

	Increasing the precision when training the ResNets with cross-entropy loss did not fully
	remove the delayed robustness to PGD, however. It only created a further delay (\figref{Resnet_adv_accs}). This is explained by skip connections causing the output variance of ResNets to scale exponentially with depth
	\citep{zhang2019fixup}, on top of naive loss minimization. The logits thus scale so fast that they reach the
	limits of both 32- and 64-bit numerical precision within a million updates, as
	shown in \figref{CE_Cifar_collapse_graph}. Batch or layer normalization is
	normally used to resolve this, as they both normalize the intermediate values after
	each skip connection. In fact, \citeauthor{ALLNETWORKSGROK} already showed an
	absence of delayed robustness to PGD when batch normalization was used in their figure 12.

	By using a stronger attack, delayed robustness was fully removed in overtrained ResNets.
	This shows that delayed robustness does not occur in larger networks with residual connections. In addition, training with normalization even removes robustness to weaker attacks.
	From this we can extrapolate that modern architectures, which use both residual connections and normalization, will not show delayed robustness when using sufficiently strong attacks.

	In contrast, we observe that in the MNIST experiments, even when we use the AA+ or ScaledPGD attack, there is still some residual robustness in networks that had loss instabilities during training.
	We have two, possibly co-occurring, explanations.
	First, it could be that this robustness is still inflated, because, while AA+ is a strong attack, it still not perfect. Other, even stronger, attacks may still be able to attack samples now deemed robust. Future work could even consider formal network verification, like with the \textit{$\alpha$,$\beta$-CROWN} verifier (see e.g. \citet{zhou2025clip}).

	The other explanation is that loss instabilities might have a small regularizing effect.
	To understand this, we remark that the loss instabilities of the MNIST experiments are reminiscent of the slingshot effect, which \citet{thilak2022slingshot} described as a non-monotonic training loss behavior characterized by sudden large spikes.
	These are especially clear when we plot the train set loss with a linear scale for the iterations, which we do in \twosupfigref{MNIST_full_res_loss}{MNIST_full_res_loss_MSE}.

	Most work on the slingshot mechanism originates from grokking literature. \citet{SOFTMAXCOLLAPSE} mention in their appendix H that the slingshot
	effect is a known side effect of adaptive gradient methods and discuss its
	relationship with grokking and softmax collapse. They state that while the
	slingshot effect can induce grokking, it is not the same as, nor a requirement
	for, grokking. This was shown by \citet{nanda2023progress}, who write in their
	own appendix (section D.2) that they suspect that slingshots \textit{``serve as
		an implicit regularization mechanism that favors the simpler, generalizing
		solution over the more complicated''}.

	We hypothesize that the slingshot mechanism is the cause of the dead neurons seen in \figref{MNIST_dead_neurons_graph}, and that after a slingshot the network's parameters move to different local loss optima that have fewer effective parameters. This new, effectively smaller network, could very well be more robust.
	To our knowledge, the relation between slingshots and dead neurons has not been studied before. Future work could determine if there truly is such a relation and, if so, whether this behavior can be exploited to benefit training, generalization, or grokking in a controlled way.

    Our work also highlights the risk of instabilities when overtraining with commonly used methods.
	Further efforts could thus also go into the design, evaluation, and
	adoption of stable versions of these algorithms. We already demonstrated the benefits of AMSGrad over Adam in this work. This is especially important if overtraining is going to be a new norm.

	Overall, delayed robustness as observed is thus primarily a result of weak adversarial attacks and training loss instabilities during overtraining. It tends to disappear with more stable training methods and stronger attacks. Grokking on the other hand, as described by for example \citet{GROKKINGPOWER} and \citet{OMNIGROK}, is often a smooth and gradual process. Furthermore, \citet{SOFTMAXCOLLAPSE} showed that higher precision benefits grokking by preventing softmax collapse, while we see that higher precision inhibits delayed robustness because it is caused by softmax collapse. We therefore conclude that delayed robustness is not grokking.

\section*{Acknowledgements}
We want to thank Humayun et al., as we appreciate that they personally helped us to reproduce their results.

\newpage
\section*{Impact Statement}
This paper critically re-assesses and refutes findings and explanations from earlier work in the field of neural networks.
We believe such scientific correction to be a valuable part of a healthy academic environment.

\bibliography{DRarxiv}
\bibliographystyle{icml2026}

\newpage
\appendix
\onecolumn
\section{Appendix}

	\subsection{Experimental Details}\label{expdetails}
	The experiments described in this paper were all performed using PyTorch
	\footnote{\url{https://pytorch.org/}} under the most basic settings possible. As
	an optimizer, we used Adam unless specified it was AMSGrad\footnote{In Pytorch
		this is still the Adam optimizer but with the setting \texttt{amsgrad = True}.},
	always with the default learning rate of $0.001$ and the default values for
	other parameters unless explicitly specified otherwise.

	The MLP networks in were all 4-layer ReLU
	MLPs with a layer width of 200. The indices of the 1000 random MNIST samples were always the same, with a batch size of 200, except in the repeat experiments mentioned in the discussion, and can also be found on our project repository \footnote{\url{https://github.com/stijnradboud/DRExplained}}.
	The ResNet18 models from \secref{resnetsubsec} were as similar as possible as
	those from \citet{ALLNETWORKSGROK}. The model-generating script we both used can
	be also found on our project repository.
	
	Like them we did not use batch-normalization and
	used a width-parameter of 16. We stored and analyzed our models on base-10 logarithmic intervals during
	training. We used a manual torch seed of 42 wherever possible before
	initializing the networks, with the exception of the repeat experiments
	mentioned in the discussion (\secref{discussion}), where we used different
	seeds.

	For the adversarial attacks, we borrowed the PGD code from
	\citet{ALLNETWORKSGROK} and it can be found on the repository, together with code for
	the ScaledPGD attack. For all attacks we used the $\epsilon$-perturbations as
	described in the main text (default $0.06$), with a step size of $0.0156$ and 10
	PGD steps. We did limit the attacks to the [0, 1]-domain. This way, copied the set-up of Humayun et al. with the exception that they used 100 PGD steps. As our results show, this does not make much difference for delayed robustness. Auto-Attack is parameter-free. The only relevant settings we used
	for that were \texttt{norm='Linf'} and \texttt{version='plus'}. We also used the $\epsilon$ specified in the main text. 
	
	Our goal with PGD was to copy the set-up of Humayun et al. and reproduce delayed robustness as reported. We chose to immediately reach for stronger attacks instead (AA+, ScaledPGD), as our main goal was to determine the adversarial robustness and not the optimal settings for PGD. We also used code from \citet{ALLNETWORKSGROK} for our local complexity results, where again we tried to replicate their set-up as close as possible. This code can be found on the repository as well.

	\subsection{Sensitivity Analysis to Learning Rate}
	We performed a sensitivity analysis the determine the effect of the learning rate in the MNIST experiments of \secref{MLPex}. In the MSE loss results, \supfigref{lrMSE}, we also compare Adam and AMSGrad and we see that a higher learning rate increases the training loss instabilities with Adam, and at the lowest learning rate the robustness to PGD even decreases. AMSGrad stabilizes the loss in all settings. The CE loss results in \supfigref{lrCE} are similar. Naive loss minimization is amplified by higher learning rates and therefore the loss instabilities occur earlier and are more extreme.

	\subsection{Training with SGD.}
	Here we provide results of using the SGD optimizer instead of Adam or AMSGrad in the MIST experiment described in figure \secref{MLPex}. We use the same training set-up but with the SGD optimizer with a learning rate of 0.01 for CE loss and 0.05 for MSE loss. We train both for 20 million iterations.
	
	The results in \twosupfigref{SGDMSE}{SGDCE} are as expected. Since SGD does not use gradient scaling, the training loss reaches very low values without any instabilities or an increase in adversarial accuracy to PGD using MSE loss. With CE loss, we do see an increasing trend in adversarial accuracy to PGD. In \supfigref{SGDSC} we show that this coincides with soft-max collapse errors, because naive loss minimization is not specific to adaptive optimizers.

	\subsection{Training with scaled cross-entropy loss.}\label{AppendixScaledCE}
	It is possible to use our ScaledSCE loss as described in \secref{SCSolutions} to also train a network. In \supfigref{scaledCE} we show results of training with ScaledSCE. All settings were similar as described in \secref{secCEMnist}, but with ScaledSCE with $s_{max}=20$ rather than default soft-max cross-entropy.
	
	The first observation is that ScaledCE does not actually inhibit Naive Loss Minimization during training: the losses when determined with normal SCE underflow to 0 and the adversarial accuracy with default PGD ends up closely matching test accuracy. This is expected since we are still using normal SCE loss but with scaled logits, thus preserving the gradient responsible for naive loss minimization. We explicitly show that NLM is preserved in \supfigref{scaledCEmag}, where we plot the magnitude of the average model output across training set samples and note that the outputs become extremely large.
	
	\subsection{Supplemental Figures}\label{supfigs}
	
	\begin{figure}[h]
		\begin{center}
			\includegraphics[width=\textwidth]{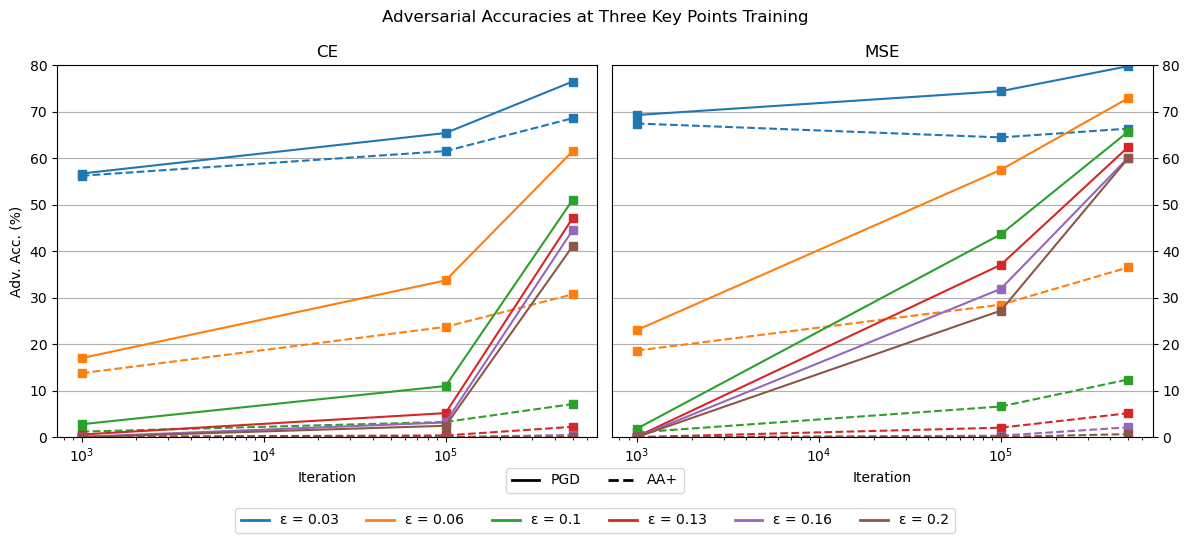}
		\end{center}
		\caption{\textit{Adversarial Accuracies when attacking the cross-entropy (left) and mean squared error (right) networks from \figref{DRfig} with both
				Auto-Attack+ and PGD at three different times during training (at iterations
				$10^3$, $10^5$ and $5 \cdot 10^5$).}}\label{PGDvsAAfig}
	\end{figure}
	\vspace{3cm}
	(Supplementary figures continue on next pages.)
	\begin{figure}[h]
		\begin{center}
			\includegraphics[width=\textwidth]{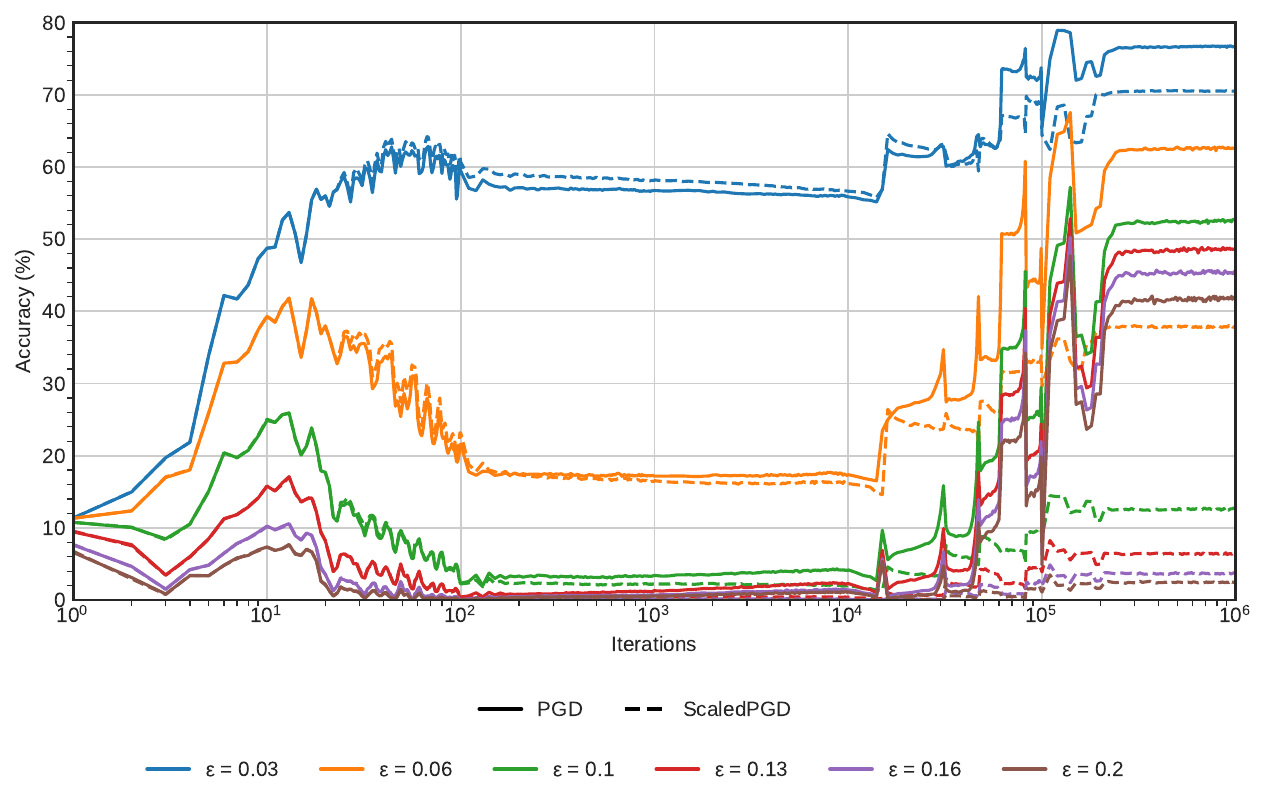}
		\end{center}
		\caption{\textit{Comparison of adversarial accuracies of the cross-entropy MLP trained on MNIST, at different $\epsilon$
				pertubation values for two attacks: $l_\infty$-PGD and $l_\infty$-ScaledPGD.}}\label{CEpgdsfig}
	\end{figure}

	\begin{figure}[h]
		\begin{center}
			\includegraphics[width=\textwidth]{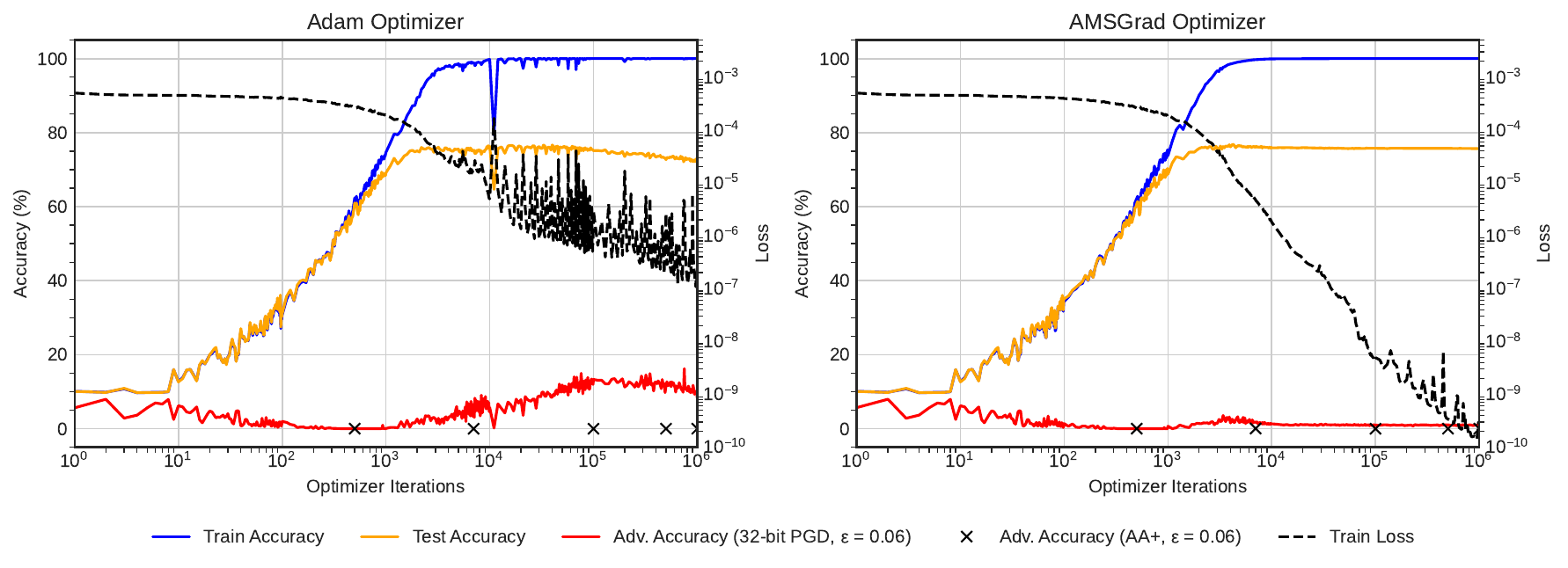}
		\end{center}
		\caption{\textit{Train, test and adversarial accuracies, from perturbing the
				test set using $l_\infty$-PGD with $\epsilon = 0.06$, as well as train set loss
				of two CIFAR10-trained ResNets with mean squared error loss. One uses default
				Adam optimizer (left), the other uses the AMSGrad optimizer
				(right).}}\label{MSECifar10Fig}
	\end{figure}

	\begin{figure}[h]
		\begin{center}
			\includegraphics[width=\textwidth]{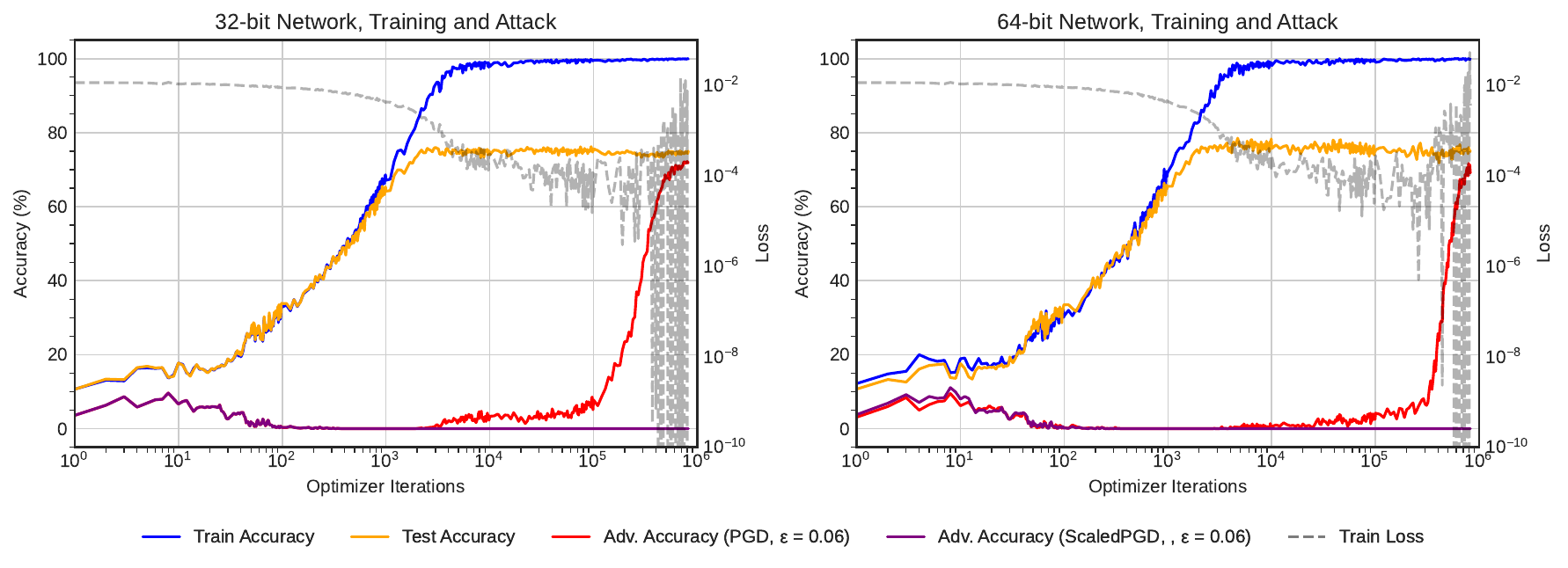}
		\end{center}
		\caption{\textit{Train, test and adversarial accuracies (from perturbing the
				test set using $l_\infty$-PGD and $l_\infty$-ScaledPGD with $\epsilon = 0.06$),
				as well as train set loss of two ResNets trained on CIFAR10 with softmax cross-entropy loss. One used a 32 bits regimen (left), the other 64 bits.}}\label{CEresnetfig}
	\end{figure}

	\begin{figure}[h]
		\begin{center}
			\includegraphics[width=\textwidth]{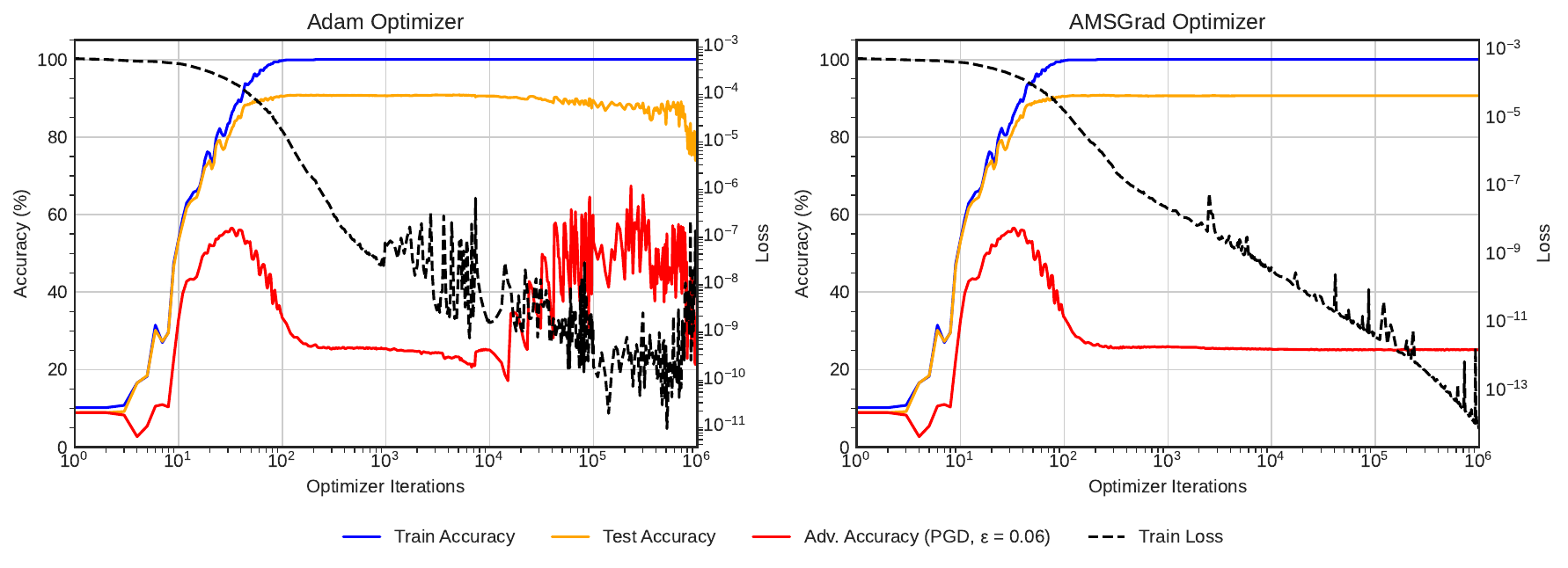}
		\end{center}
		\caption{\textit{Train, test and adversarial accuracies (from perturbing the
				test set using $l_\infty$-PGD with $\epsilon = 0.06$), as well as train set loss
				of two MNIST-trained MLPs trained with mean squared error loss and
				\textbf{leakyReLU activation functions}. One uses the default Adam optimizer
				(left) the other the AMSGrad optimizer (right).}}\label{MSEMnistfigleaky}
	\end{figure}

	\begin{figure}[h]
		\begin{center}
			\includegraphics[width=\textwidth]{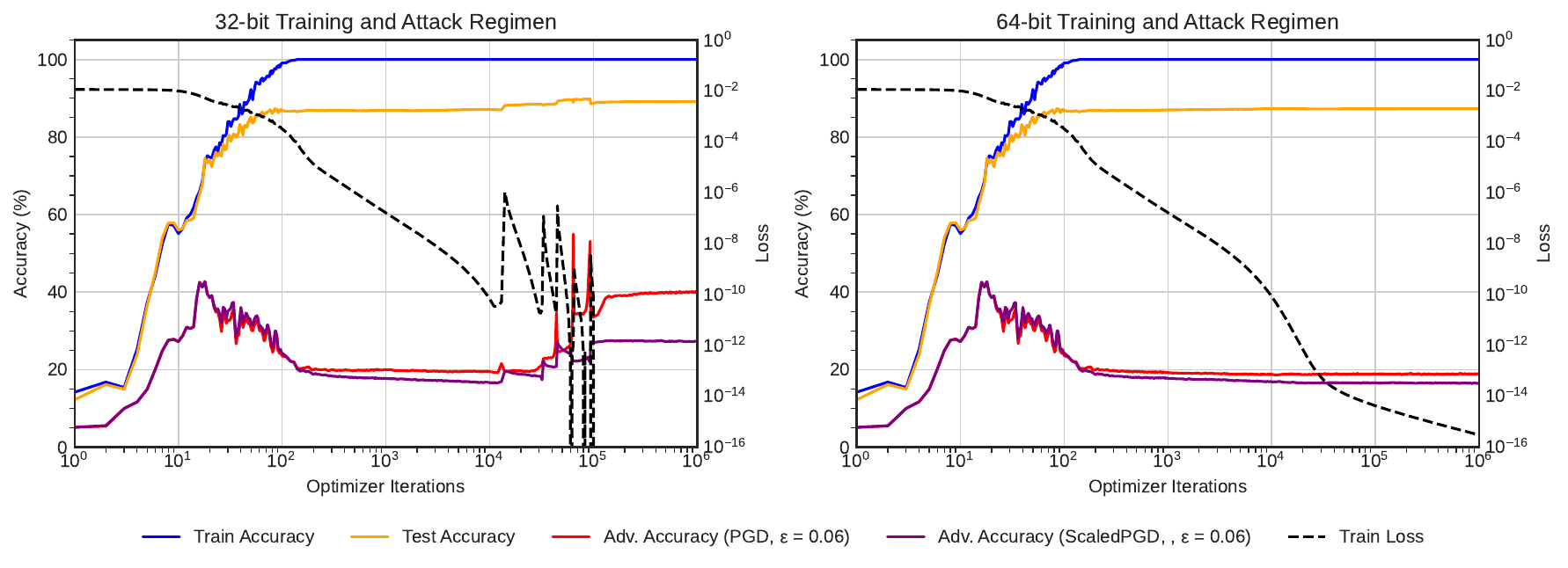}
		\end{center}
		\caption{\textit{Train, test and adversarial accuracies (from perturbing the
				test set using $l_\infty$-PGD and
				$l_\infty$-ScaledPGD) with $\epsilon = 0.06$), as well as train set loss of two
				MNIST-trained MLPs trained with softmax cross-entropy loss and
				\textbf{leakyReLU activation functions}. One was trained using default 32-bit
				loss calculation (left), the other used a 64-bit loss
				calculation.}}\label{CEMnistfigleaky}
	\end{figure}

	\begin{figure}[h]
		\begin{center}
			\includegraphics[width=\textwidth]{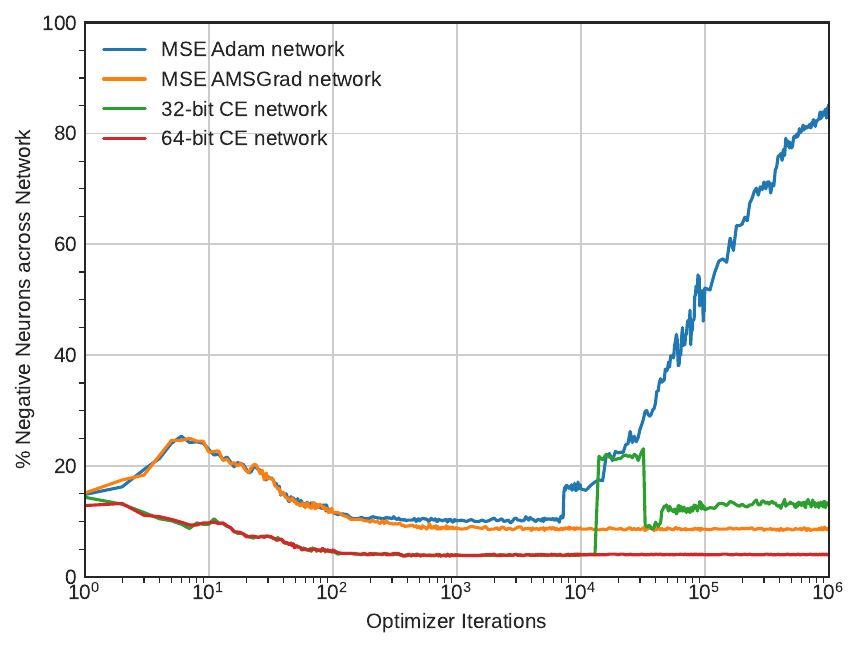}
		\end{center}
		\caption{\textit{Analysis of the number of fully negative neurons throughout
				training of the MLPs \textbf{using leakyReLU activation functions} on 1000 MNIST
				samples in our experimental set-ups.}}\label{MNIST_dead_neurons_graph_leaky}
	\end{figure}

	\begin{figure}[h]
		\begin{center}
			\includegraphics[width=\textwidth]{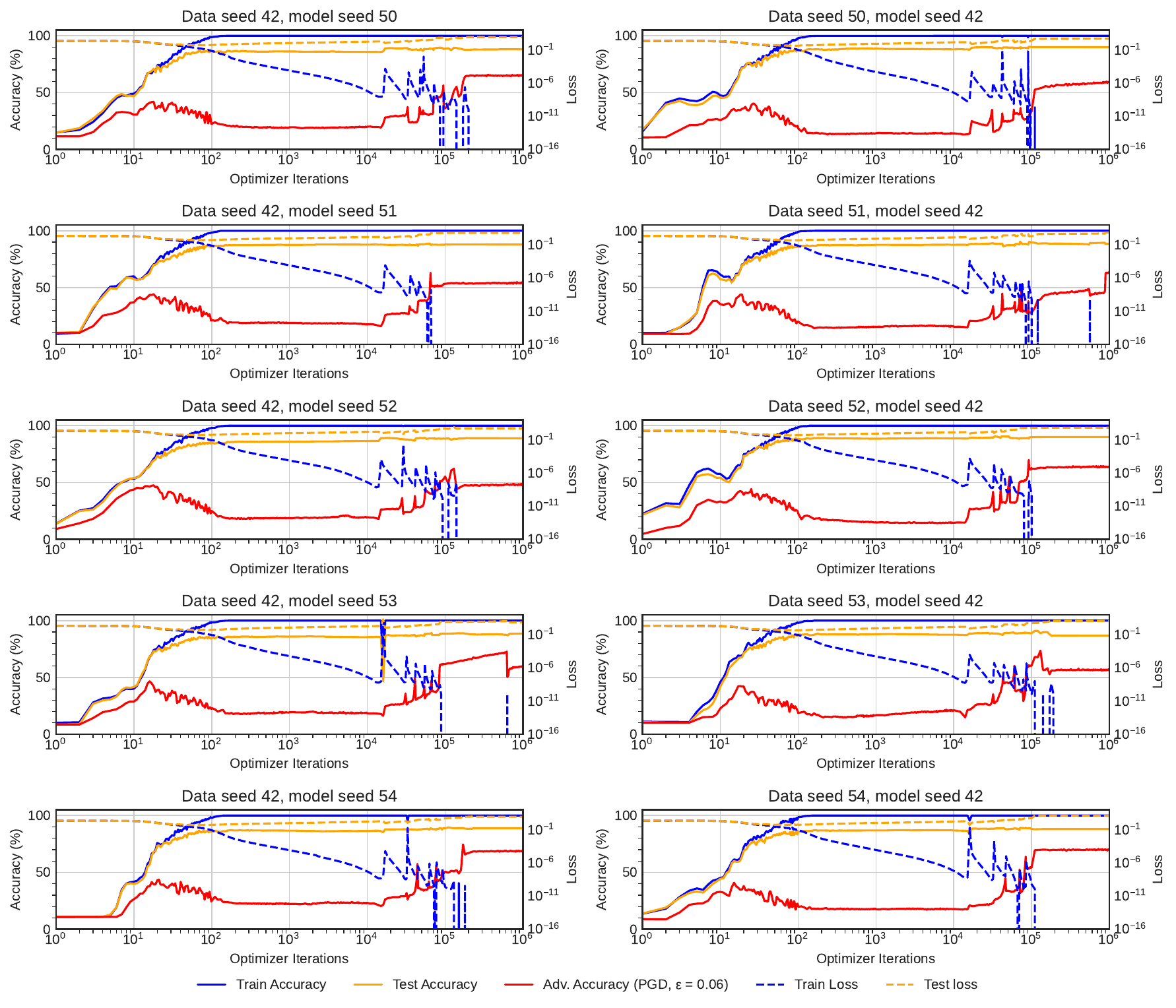}
		\end{center}
		\caption{\textit{Accuracy and loss results for 10 repeat runs with different
				sampling- and model seeds for the MNIST MLP
				experiments using CE loss.}}\label{MNIST_repeat_accs_CE}
	\end{figure}

	\begin{figure}[h]
		\begin{center}
			\includegraphics[width=\textwidth]{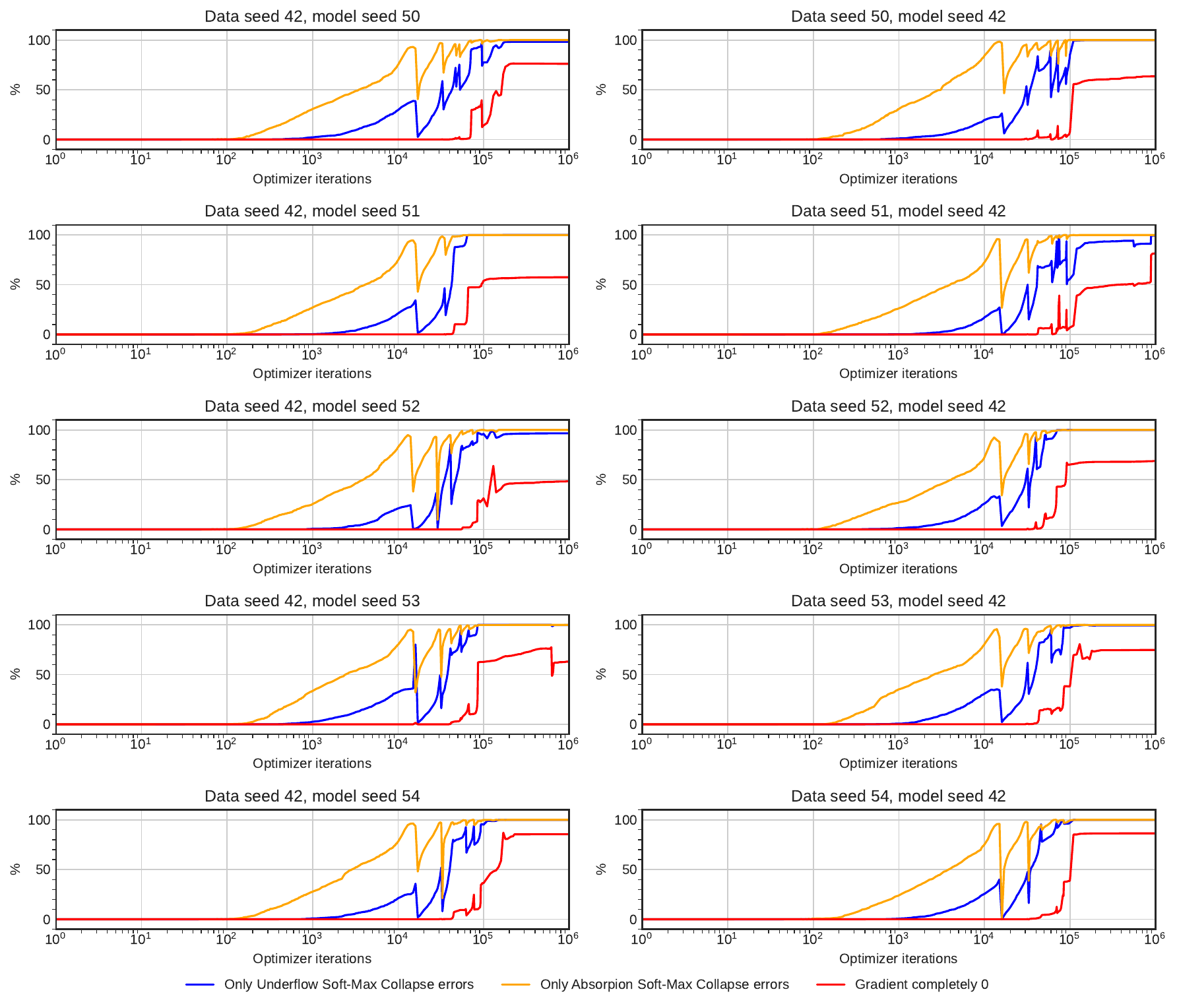}
		\end{center}
		\caption{\textit{Softmax collapse results for 10 repeat runs with
				different sampling- and model seeds for the MNIST MLP
				experiments using CE loss.}}\label{MNIST_repeat_grads_CE}
	\end{figure}
	
	\begin{figure}[h]
		\begin{center}
			\includegraphics[width=\textwidth]{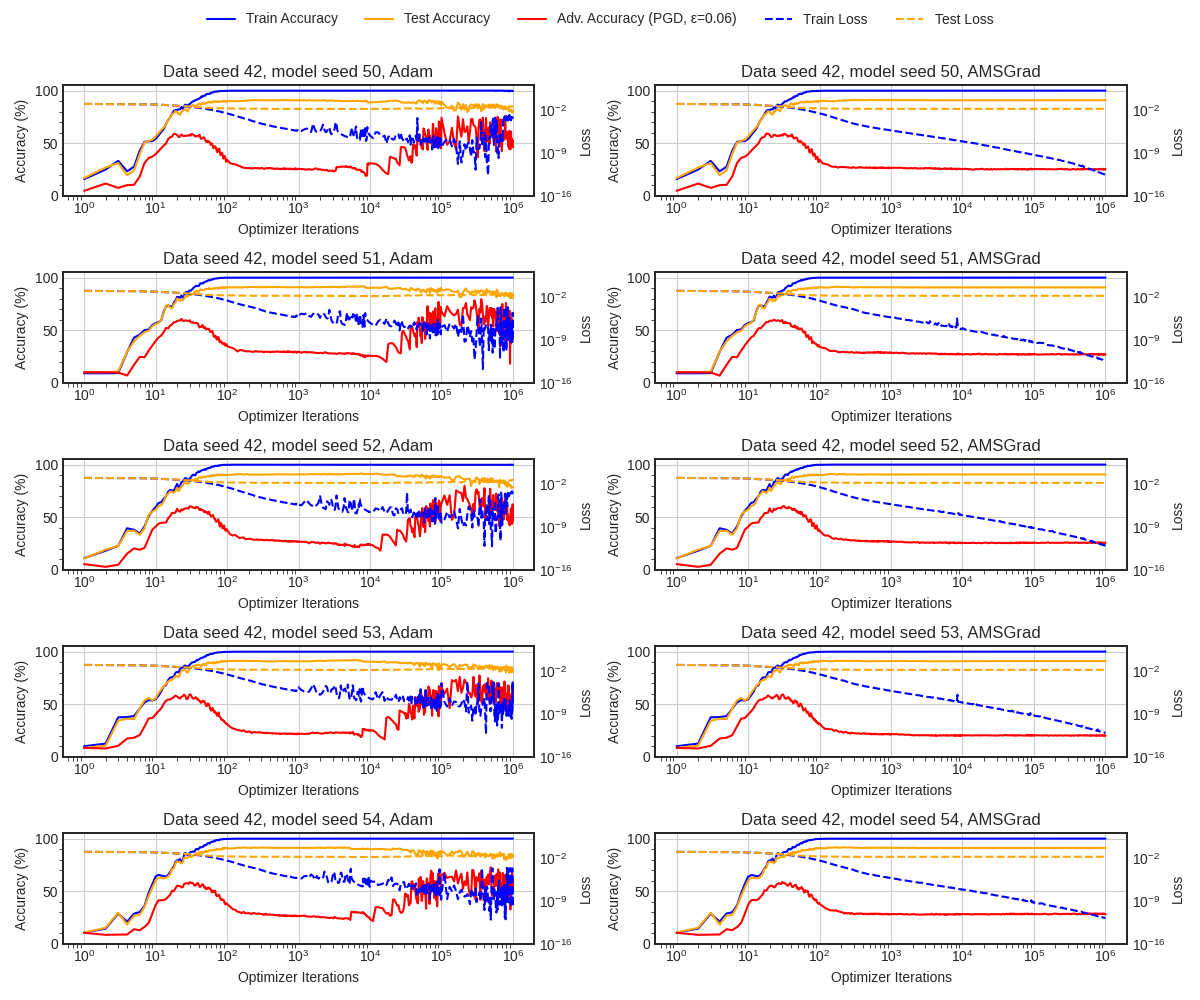}
		\end{center}
		\caption{\textit{Accuracy and loss results for 5 repeat runs with different
				model seeds for the MNIST MLP
				experiments using MSE loss.}}\label{MNIST_repeat_accs_model_MSE}
	\end{figure}
	
	\begin{figure}[h]
		\begin{center}
			\includegraphics[width=\textwidth]{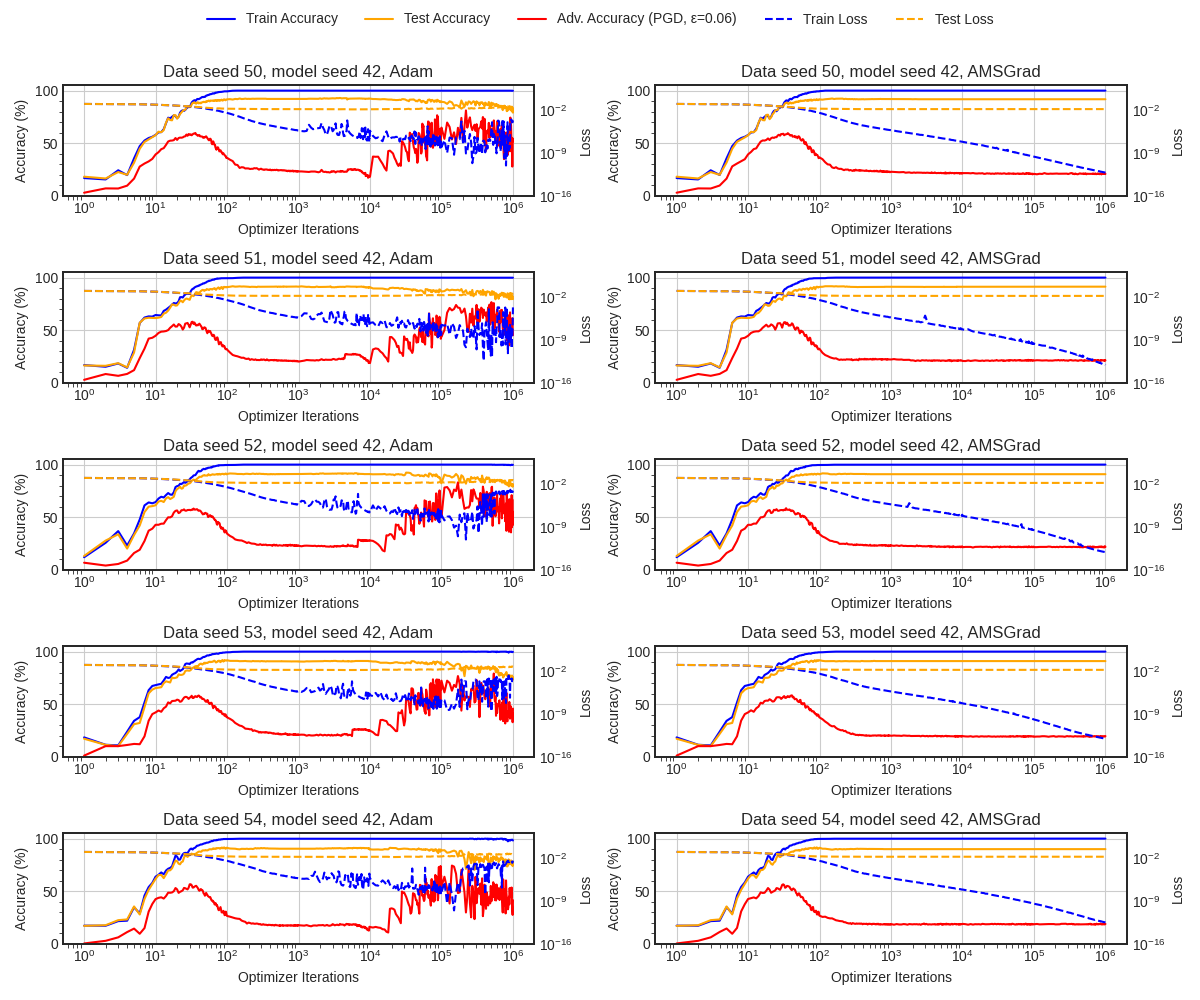}
		\end{center}
		\caption{\textit{Accuracy and loss results for 5 repeat runs with different
				data subsample seeds for the MNIST MLP
				experiments using MSE loss.}}\label{MNIST_repeat_accs_data_MSE}
	\end{figure}

	\begin{figure}[h]
		\begin{center}
			\includegraphics[width=\textwidth]{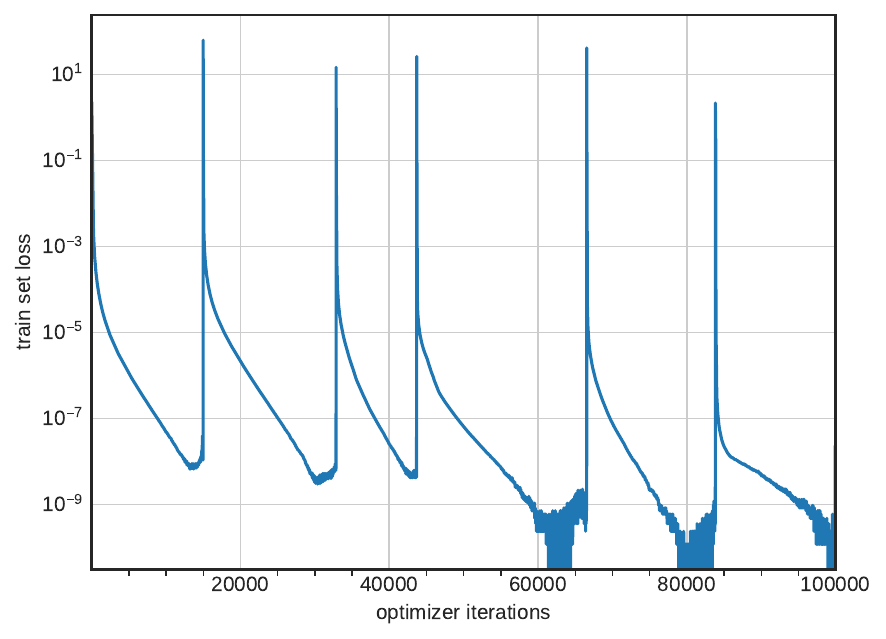}
		\end{center}
		\caption{\textit{The train set \textbf{cross-entropy} loss of the first 100k iterations of a repeat
				experiment similar to \figref{MNIST_repeat_accs_CE} (model seed 42, data seed 42)
				to show slingshot effect-like behavior. Contrary to other plots
				in this paper, these were recorded at every iteration and the x-axis is \textbf{not}
				log-scaled.}}\label{MNIST_full_res_loss}
	\end{figure}

	\begin{figure}[h]
		\begin{center}
			\includegraphics[width=\textwidth]{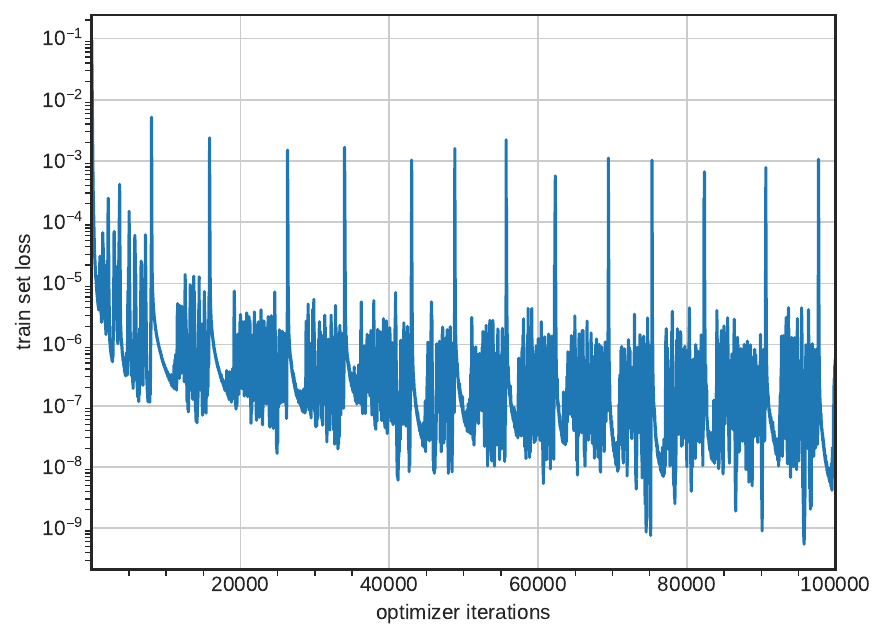}
		\end{center}
		\caption{\textit{The train set \textbf{mean squared error} loss of the first 100k iterations of a repeat
				experiment similar to \figref{MNIST_repeat_accs_model_MSE} (model seed 42, data seed 42)
				to properly show slinghot effect-like behavior. Contrary to other plots in
				this paper, these were recorded at every iteration and the x-axis is \textbf{not}
				log-scaled.}}\label{MNIST_full_res_loss_MSE}
	\end{figure}
	
	\begin{figure}[h]
		\begin{center}
			\includegraphics[width=\textwidth]{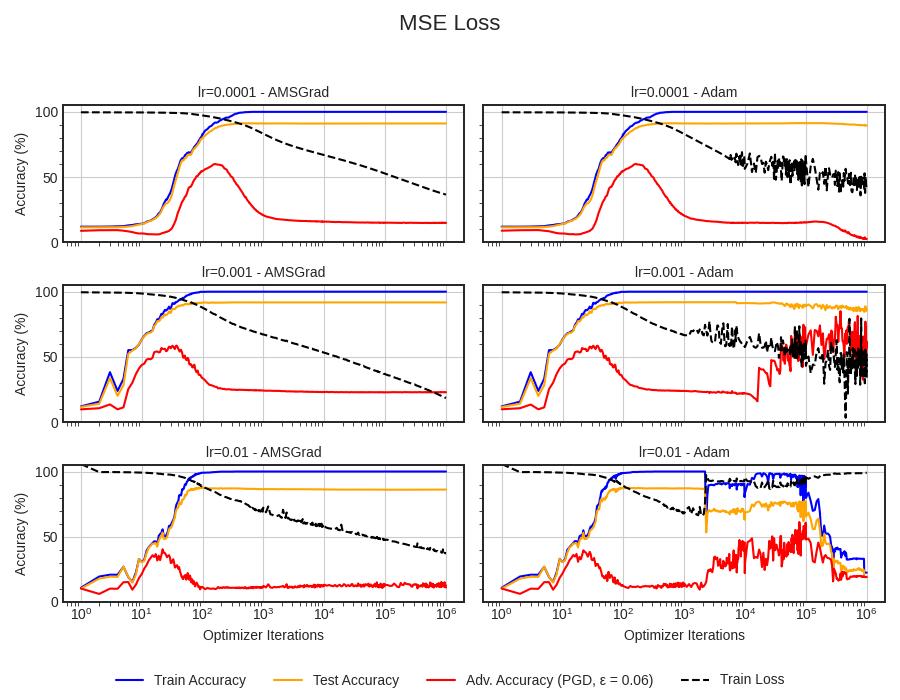}
		\end{center}
		\caption{\textit{The effects of varying the learning rate with the different optimizers using MSE Loss. Note that 0.001 is the default.}}\label{lrMSE}
	\end{figure}
	\begin{figure}[h]
		\begin{center}
			\includegraphics[width=\textwidth]{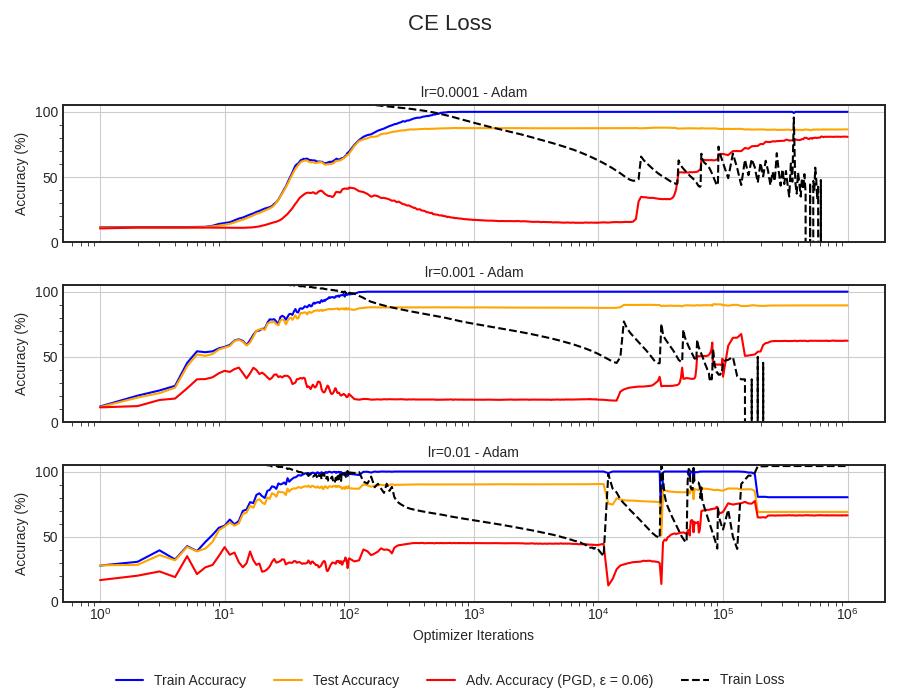}
		\end{center}
		\caption{\textit{The effects of varying the learning rate using CE Loss. Note that 0.001 is the default.}}\label{lrCE}
	\end{figure}
	
	\begin{figure}[h]
		\begin{center}
			\includegraphics[width=0.5\linewidth]{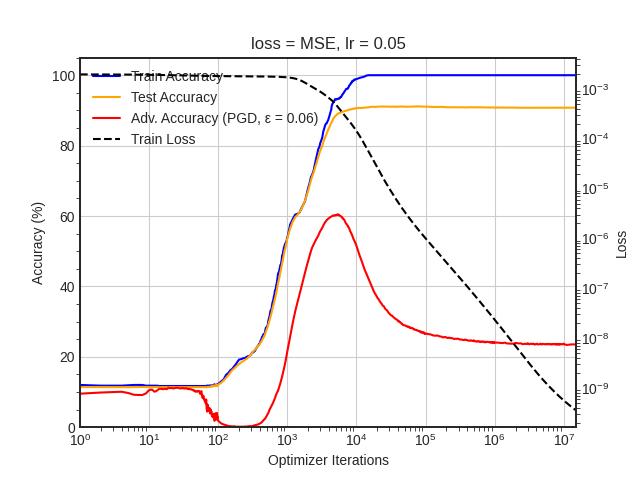}
		\end{center}
		\caption{\textit{Train, test and adversarial accuracies (from perturbing the
				test set using $l_\infty$-PGD with and AA+ $\epsilon = 0.06$) of an MNIST-trained MLP
				trained with mean squared error loss with the SGD optimizer.}}\label{SGDMSE}
	\end{figure}
	
	\begin{figure}[h]
		\begin{center}
			\includegraphics[width=0.5\linewidth]{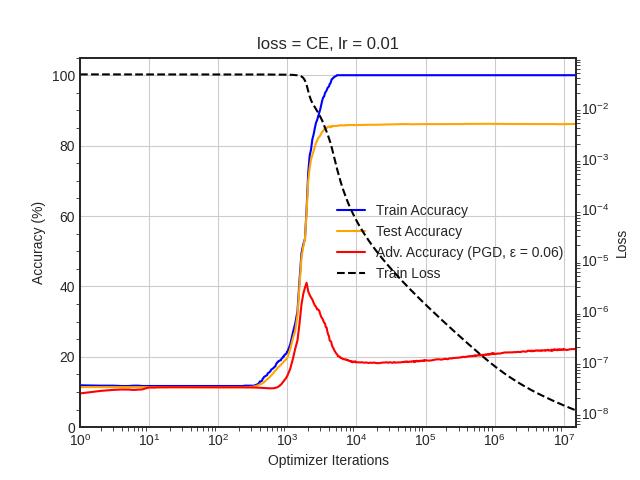}
		\end{center}
		\caption{\textit{Train, test and adversarial accuracies (from perturbing the
				test set using $l_\infty$-PGD with and AA+ $\epsilon = 0.06$) of an MNIST-trained MLP
				trained with soft-max cross-entropy loss with the SGD optimizer.}}\label{SGDCE}
	\end{figure}
	
	\begin{figure}[h]
		\begin{center}
			\includegraphics[width=0.5\linewidth]{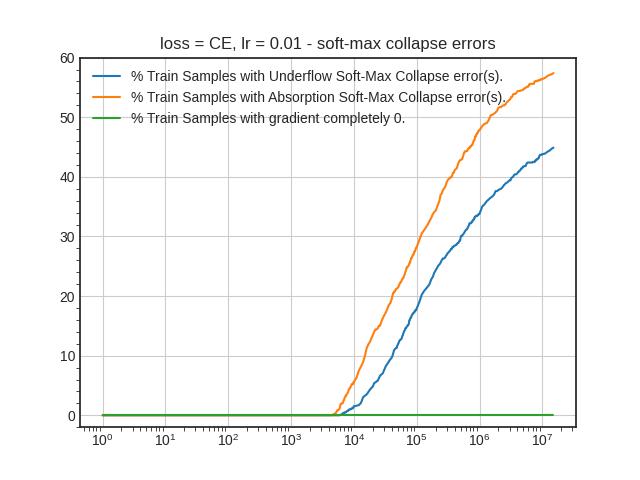}
		\end{center}
		\caption{\textit{Soft-max collapse errors on the train set samples in an MNIST-trained MLP
				trained with soft-max cross-entropy loss with the SGD optimizer.}}\label{SGDSC}
	\end{figure}
	
	\begin{figure}[h]
		\begin{center}
			\includegraphics[width=0.5\textwidth]{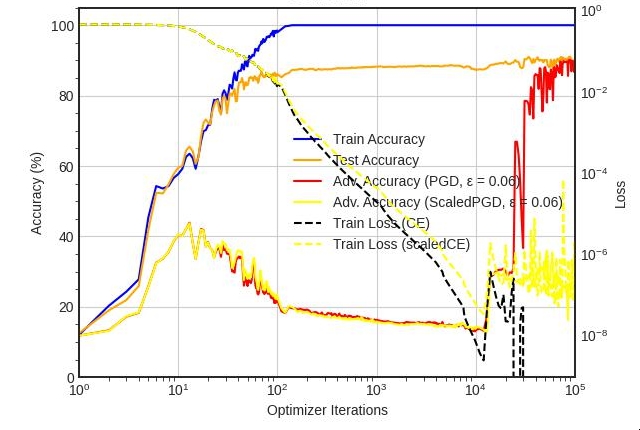}
		\end{center}
		\caption{\textit{Results of training with scaled SCE and Adam. Note that the train loss for standard soft-max cross-entropy was only included for comparison and not used to actually train the network.}}\label{scaledCE}
	\end{figure}
	
	\begin{figure}[h]
		\begin{center}
			\includegraphics[width=0.75\textwidth]{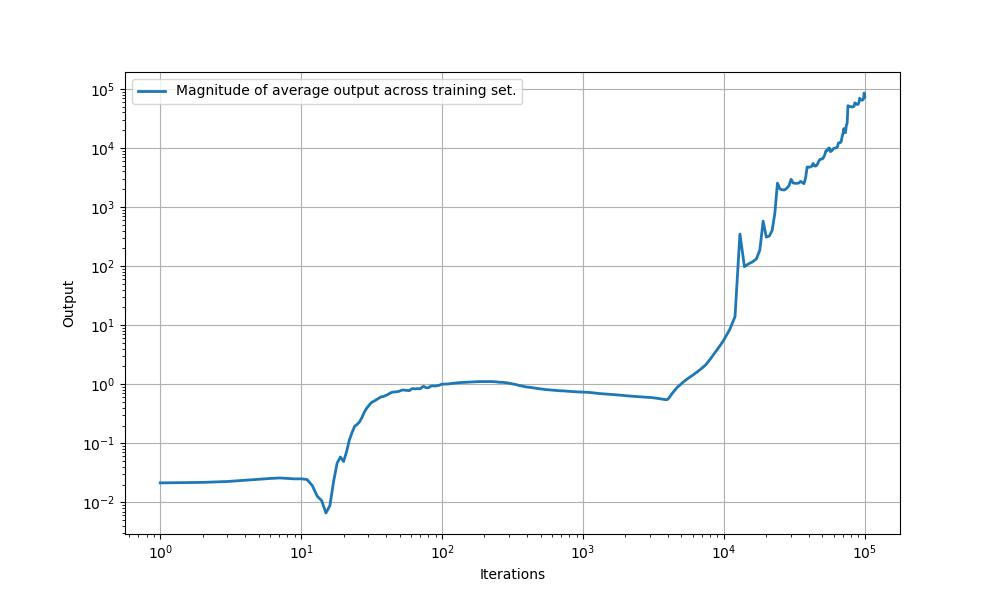}
		\end{center}
		\caption{\textit{Magnitude of the average output across training samples of the network trained with ScaledSCE.}}\label{scaledCEmag}
	\end{figure}

\end{document}